%% file: main.tex
\documentclass[10pt,twocolumn,letterpaper]{article}

\usepackage[pagenumbers]{cvpr} 

\usepackage{graphicx}
\usepackage{amsmath}
\usepackage{amssymb}
\usepackage{booktabs}
\usepackage{multirow}
\usepackage{algorithm}
\usepackage{algorithmic}
\usepackage{wrapfig}
\usepackage{xcolor}
\usepackage{enumitem}
\setitemize{itemsep=10pt,topsep=0pt,parsep=0pt,partopsep=0pt}

\usepackage{hyperref}
\hypersetup{pagebackref,breaklinks,colorlinks}

\usepackage[capitalize]{cleveref}
\crefname{section}{Sec.}{Secs.}
\Crefname{section}{Section}{Sections}
\Crefname{table}{Table}{Tables}
\crefname{table}{Tab.}{Tabs.}

\def\cvprPaperID{*****} 
\def\confName{CVPR}
\def\confYear{2023}

\definecolor{darkred}{rgb}{0.7,0,0}

\usepackage{color, colortbl}
\definecolor{Gray}{gray}{0.93}
\definecolor{orange}{rgb}{0.9,0.5,0}
\definecolor{beaublue}{rgb}{0.74, 0.83, 0.9}
\definecolor{unitednationsblue}{rgb}{0.36, 0.57, 0.9}

\newcommand{\shortname}{\textsc{Lafite2}}
\newcommand{\shortnamecvpr}{\textsc{Lafite}}

\newcommand{\lafitegan}{\textsc{Lafite2}$_{\textbf{\texttt{GAN}}}$}
\newcommand{\lafiteldm}{\textsc{Lafite2}$_{\textbf{\texttt{LDM}}}$}

\input{definitions.tex}

\title{\shortname{}: Few-shot Text-to-Image Generation}

\begin{document}


\author{Yufan Zhou\\
State University of New York at Buffalo\\
{\tt\small yufanzho@buffalo.edu}
\and
Chunyuan Li\\
Microsoft\\
{\tt\small chunyl@microsoft.com}
\and
Changyou Chen\\
State University of New York at Buffalo\\
{\tt\small changyou@buffalo.edu}
\and
Jianfeng Gao\\
Microsoft\\
{\tt\small jfgao@microsoft.com}
\and
Jinhui Xu\\
State University of New York at Buffalo\\
{\tt\small jinhui@buffalo.edu}
}
\maketitle


\begin{abstract}
    Text-to-image generation models have progressed considerably in recent years, which can now generate impressive realistic images from arbitrary text. Most of such models are trained on web-scale image-text paired datasets, which may not be affordable for many researchers. In this paper, we propose a novel method for pre-training text-to-image generation model on image-only datasets. It considers a retrieval-then-optimization procedure to synthesize pseudo text features: for a given image, relevant pseudo text features are first retrieved, then optimized for better alignment. The low requirement of the proposed method yields high flexibility and usability: it can be beneficial to a wide range of settings, including the few-shot, semi-supervised and fully-supervised learning; it can be applied on different models including generative adversarial networks (GANs) and diffusion models. Extensive experiments illustrate the effectiveness of the proposed method. On MS-COCO dataset, our GAN model obtains Fr\'echet Inception Distance (FID) of 6.78 which is the new state-of-the-art (SoTA) of GANs under fully-supervised setting. Our diffusion model obtains FID of 8.42 and 4.28 on zero-shot and supervised setting respectively, which are competitive to SoTA diffusion models with a much smaller model size.
\end{abstract}
\section{Introduction}
Text-to-image (T2I) generation is an appealing research topic in computer vision, due to its flexibility allowed by using natural languages as the instructions, and high fidelity in the generated images in recent years. General-purpose text-to-image generation is still quite challenging because of the difficulty in bridging different modalities. 
To address such an issue, a considerable amount of effort has been devoted to this problem, and 
significant progresses have already been made  
by 
designing novel gigantic generative models and training them on web-scale image-text paired datasets. Starting from DALL-E~\cite{ramesh2021zero}, it becomes possible to perform zero-shot text-to-image generation with arbitrary text input. It quickly inspires several follow-up works, including CogView~\cite{ding2021cogview}, latent diffusion model (LDM)~\cite{rombach2021high}, GLIDE~\cite{nichol2021glide} and DALL-E 2~\cite{ramesh2022hierarchical}. These works further improved the results in terms of both human judgement and quantitative evaluation, by incorporating diffusion models~\cite{dhariwal2021diffusion,ho2020denoising} into model designs, which achieved impressive results in other generation tasks. 

The aforementioned methods, however, also exhibit their own drawbacks, preventing them from being widely used. 
For example, these models typically consist of billions of parameters, and thus require training on hundreds of millions of image-text pairs. The high computational cost and the demand of high-quality web-scale datasets could be major obstacles for most research teams in the community. 
Furthermore, despite their impressive ability in handling arbitrary text, the {\bf few-shot transferability} of these models has been less explored. We aruge that this is actually a quite important ability in practice, especially when one wants to generate image samples of certain styles or in a customized domain with limited in-domain image-text paired samples accessible. In the literature, it is known that training and fine-tuning large unconditional generative models with limited samples is not an easy task~\cite{karras2020training,zhao2020leveraging,zhao2020differentiable,ojha2021few}, not to mention that one needs to
additionally consider image-text correspondence besides image quality. This makes few-shot text-to-image generation much more challenging than unconditional generation.

Different from these models, \shortnamecvpr{}~\cite{zhou2021lafite} and KNN-Diffusion~\cite{ashual2022knn} investigate a new direction, i.e., unsupervised or the so-called \textit{language-free} methods. Distinctive from the aforementioned generative models that require training on image-text pairs, they propose to train text-to-image generation models with only image samples by utilizing the multi-modal feature space of CLIP~\cite{radford2021learning}. 
Because CLIP is trained to encode image and text into the same multi-modal feature space, a model conditioned on CLIP image features can also perform inference with CLIP text features. 
These methods are much more affordable because image samples can be easily collected, while high-quality image-text pairs need considerably more extra efforts including human captioning and filtering. However, the performances of these language-free models are still not satisfactory compared to fully-supervised methods and the aforementioned large models.

In this work, we further exploit the unsupervised text-to-image generation model, and advocate that the framework is an excellent fit for few-shot conditional generation. Starting from analyzing the performance gap between previous language-free and fully-supervised methods, we propose a new language-free method that obtains new state of the arts. 

\begin{figure*}[ht!]
    \begin{subfigure}{0.23\linewidth}
        \includegraphics[width=1.\linewidth]{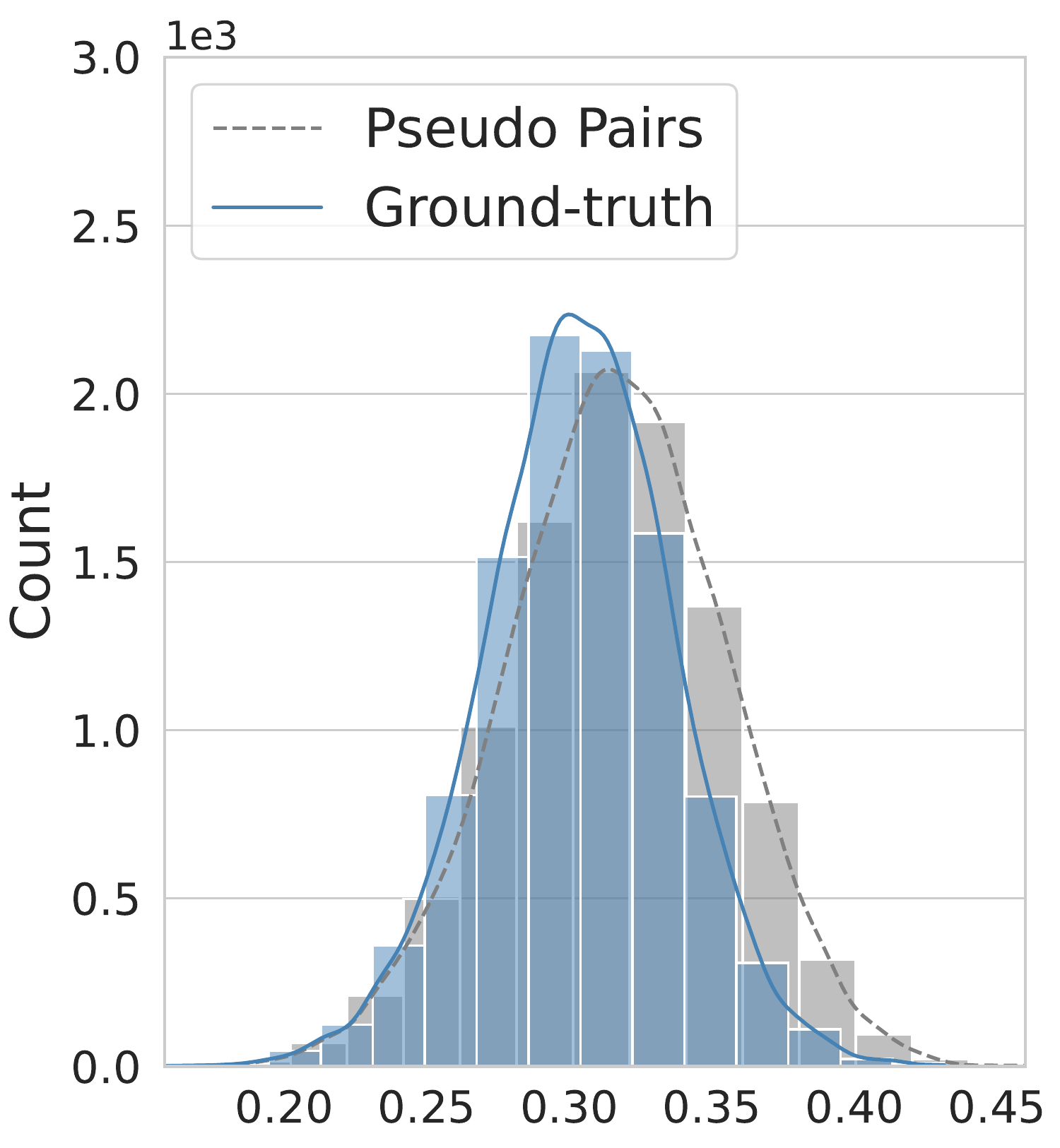}
        \caption{Image-text}
    \end{subfigure}
    \begin{subfigure}{0.23\linewidth}
        \includegraphics[width=1.\linewidth]{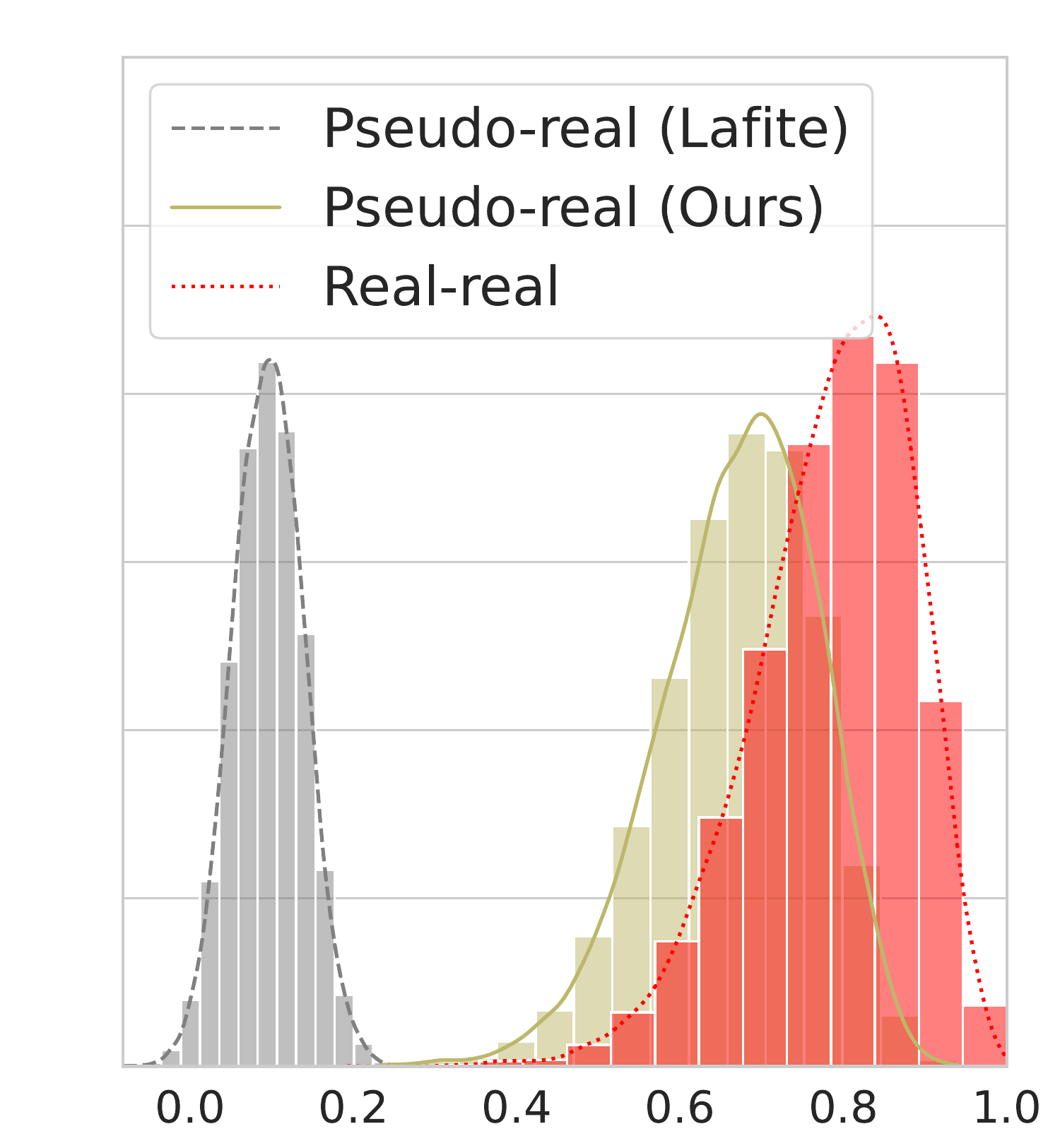}
        \caption{Text-text (paired)}
    \end{subfigure}
    \begin{subfigure}{0.23\linewidth}
        \includegraphics[width=1.\linewidth]{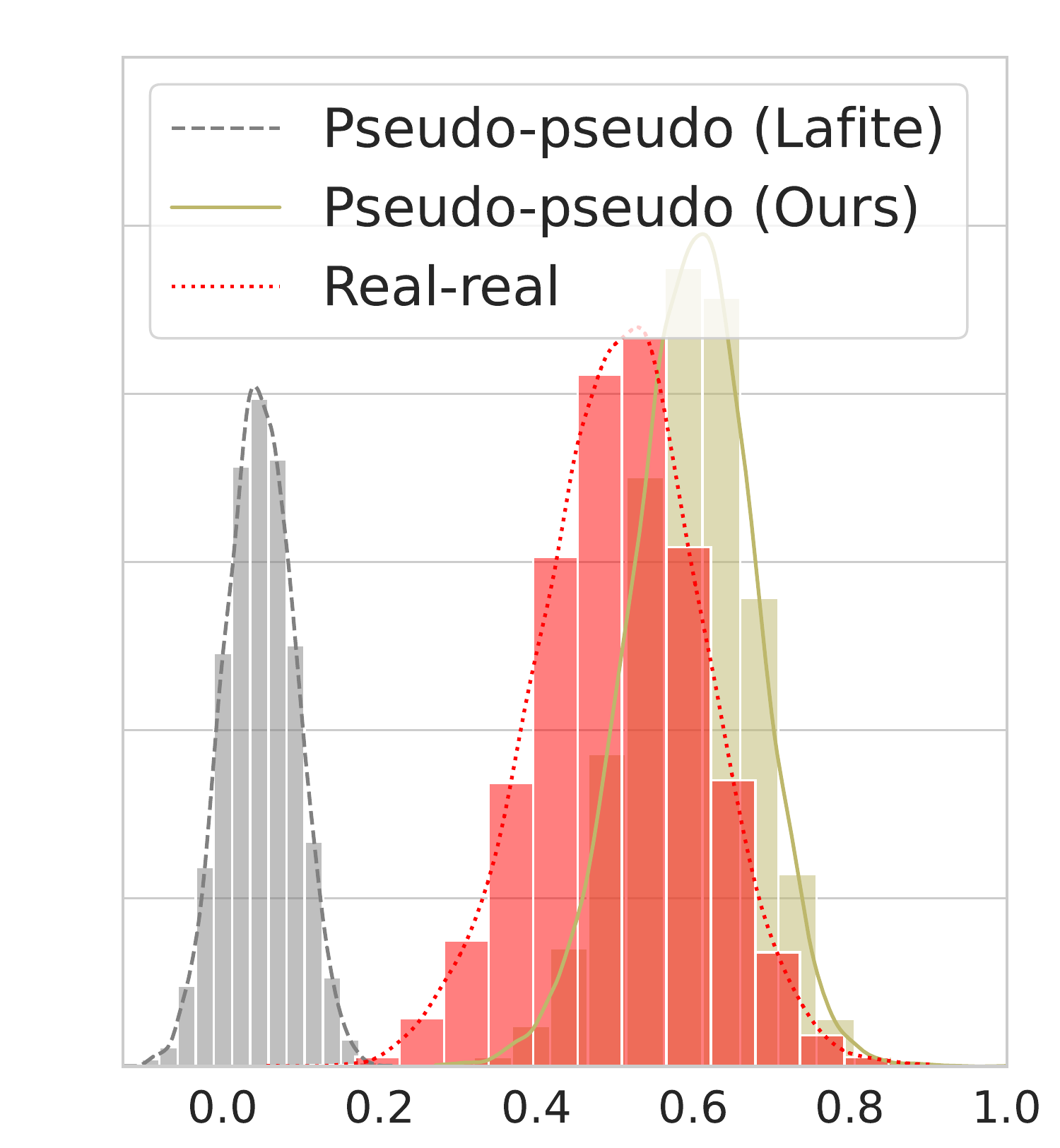}
        \caption{Text-text (unpaired)}
    \end{subfigure}
    \begin{subfigure}{0.23\linewidth}
        \includegraphics[width=1.\linewidth]{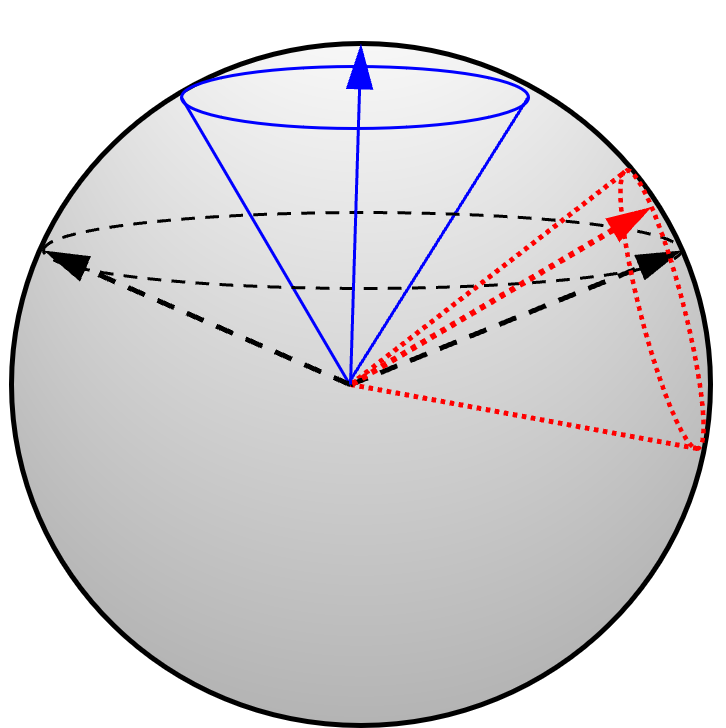}
        \caption{CLIP feature space}
    \end{subfigure}    
    \caption{Distributions of cosine similarities for (a) image-text pair (including pseudo and ground-truth text); (b) text-text pairs from the same image; (c) text-text pairs from randomly sampled images. (d) Illustration of multi-modal feature space of CLIP, where \textcolor{blue}{blue solid arrow} denotes an image feature, \textcolor{red}{red dotted arrow} denotes text feature of corresponding ground-truth caption, black dashed arrows denote possible pseudo text features generated with \shortnamecvpr{}.}
    \label{fig:similarity}
    \vspace{-0.2in}
\end{figure*}
Our main contributions can be summarized as follows:
\begin{itemize}[leftmargin=5mm]
    \setlength\itemsep{0em}
    \item We propose \shortname{} for language-free text-to-image generation, consisting of two novel techniques to synthesize pseudo text features: retrieval-augmented pseudo text feature construction, and latent feature optimization.
    
    \item We apply the proposed techniques to two widely used T2I model families, respectively. It leads to two models: 
    $(i)$ \lafitegan{}  when applied to GANs-based methods such as \shortnamecvpr{}~\cite{zhou2021lafite} and 
    $(ii)$ \lafiteldm{}  when applied to diffusion-based methods such as LDM~\cite{rombach2021high}. \shortname{} demonstrates good transferability in a wide range scenarios, benefiting few-shot, semi-supervised and fully-supervised text-to-image generation;
    
    
    
    \item We show the proposed \shortname{} obtains new SoTA results of GANs on fully-supervised text-to-image generation, with a Frechet Inception Distance (FID) score of 6.78 on MS-COCO~\cite{lin2014microsoft}. \shortname{} also obtains new SoTA results in the language-free setting. Furthermore, it also demonstrates comparable zero-shot results in relative to the recent SoTA achieved on particularly large models.
\end{itemize}

\section{Preliminaries: Probing Multimodal Feature Space}

Through generating pseudo image-text pairs in the multi-modal joint feature space of CLIP, \shortnamecvpr{} \cite{zhou2021lafite} is able to learn image-text correspondences from image-only data and perform text-to-image generation. However, the quantitative results of training on pseudo image-text pairs are much worse than training on ground-truth image-text pairs as shown in \cite{zhou2021lafite}. We suspect that this happens because part of the generated fake text features might be too distinctive from text features of real captions.

To verify this assumption, we conduct an ablation study. We randomly sample 10,000 images with their associated 50,000 captions from MS-COCO~\cite{lin2014microsoft} dataset, and construct pseudo text features $\{\hb_{ij}\}$ for image $\xb_i$, by following the method in \cite{zhou2021lafite}, as:
\begin{align}\label{eq:lafite}
    \hb_{ij} &= \tilde{\hb_{ij}} / \Vert \tilde{\hb_{ij}} \Vert_2, \nonumber \\
    \tilde{\hb_{ij}} &= f_{\text{img}}(\xb_i) + \xi \Vert f_{\text{img}}(\xb_i) \Vert_2 \epsilonb_{ij}/\Vert \epsilonb_{ij} \Vert_2, ~\epsilonb_{ij} \sim \mathcal{N}(\mathbf{0}, \mathbf{I}).
\end{align}
where $f_{\text{img}}$ dentoes the image encoder of CLIP, subscript $j$ is used to distinguish different pseudo features for the same image $\xb_i$. 
We compute image-text and text-text cosine similarities inside the multi-modal feature space of CLIP, whose distributions are plotted in Figure \ref{fig:similarity}. 
Although similarity distributions of pseudo pairs and ground-truth pairs heavily overlap, Figure \ref{fig:similarity} shows that pseudo text features generated by \shortnamecvpr{} are quite different from the real ones. 
From the similarity of text features associated with different images, we can also see that the effective output space of CLIP text encoder is very small, while the pseudo text features occupy a much larger region. A recent work \cite{Liang2022MindTG} also reveals a related phenomenon in the feature spaces of multi-modal models like CLIP: features from different encoders concentrate on different narrow cones of the feature space, and different modalities are actually not well-aligned in the multi-modal feature space. 

We now explain how the above observations influence the model performance in the text-to-image generation task. 
We use $\mathcal{I}$ and $\mathcal{T}$ to denote two effective output spaces, which are sets of all possible outputs from the image and text encoders, respectively, visualized in Figure \ref{fig:similarity}. We use spherical cone with blue solid lines to represent $\mathcal{I}$, while the red dotted lines corresponding to $\mathcal{T}$. The set of all possible pseudo text features generated with \eqref{eq:lafite} is denoted as  $\mathcal{H}_{\xi}$ and represented with black dashed lines.  $\mathcal{H}_{\xi}$ is constructed by adding noises to image features as \eqref{eq:lafite}, thus $\mathcal{I} \subset \mathcal{H}_{\xi}$.
Following~\cite{Liang2022MindTG}, we do not assume overlapping between $\mathcal{I}$ and $\mathcal{T}$. However, our conclusion to be presented later still holds even if they do overlap.


To ensure that  the model can learn image-text correspondence, $\mathcal{H}_{\xi}$ must overlap with $\mathcal{T}$ so that some generated pseudo text features lie within $\mathcal{T}$. According to \eqref{eq:lafite}, $\xi$ has to be large enough to enable the overlapping. From~\cite{zhou2021lafite}, \shortnamecvpr{} achieves best results with $\xi=3$, meaning that huge noises have been injected.
However, such large perturbation can also generates many fake text features that are outside of $\mathcal{T}$. Because all the real text features at inference time will always be in $\mathcal{T}$, these fake features outside of $\mathcal{T}$ simply behave as outliers, which can significantly degenerate the testing performance. Consequently, we aim to avoid generating these noisy features. To this end, instead of generating text features by directly perturbing image features as \eqref{eq:lafite}, we propose a new method to ensure that the generated fake text features are well located inside of $\mathcal{T}$. 

\section{Proposed Method: Retrieval-then-Optimization}

\begin{figure*}[t!]
    \centering
    \begin{subfigure}{0.59\linewidth}
        \includegraphics[width=0.99\linewidth]{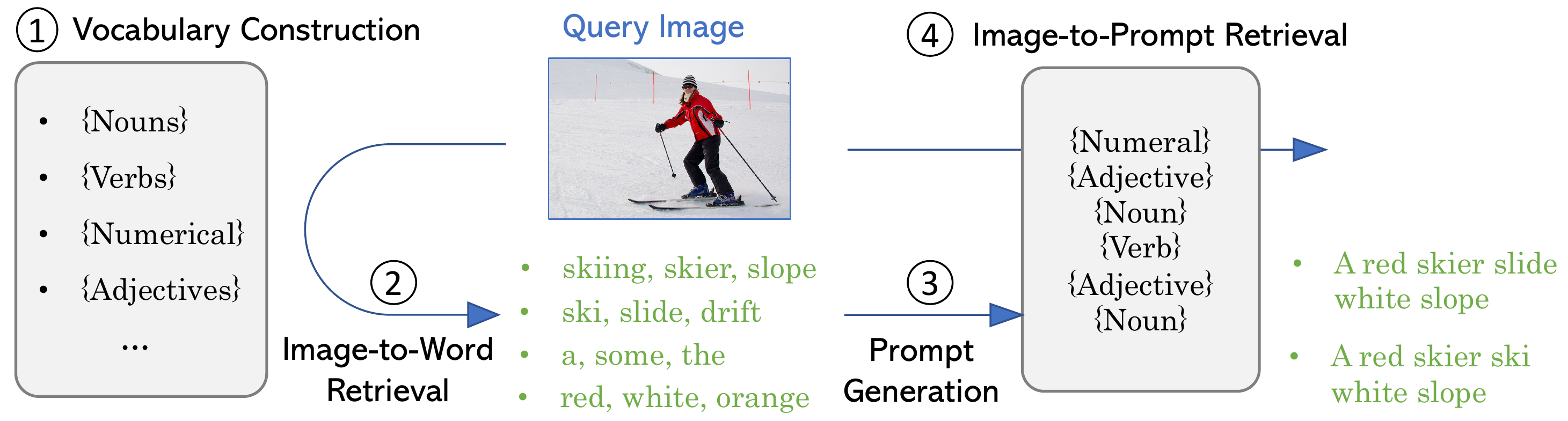}
        \caption{Retrieval-augmented synthesis}
    \end{subfigure}
    \begin{subfigure}{0.30\linewidth}
        \includegraphics[width=0.9\linewidth]{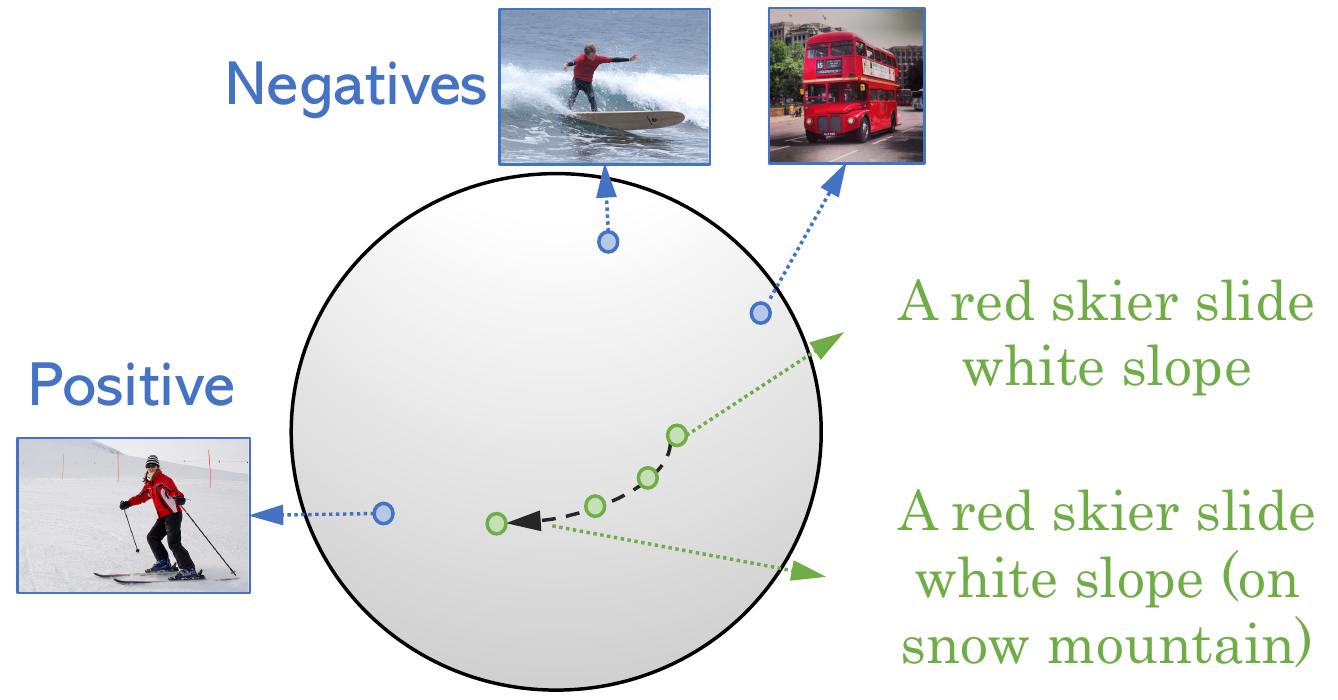}
        \caption{Contrastive latent optimization}
    \end{subfigure}    
    \caption{Illustration of the proposed method, which first generates pseudo text features and optimizes them inside of the multi-modal feature space of CLIP. In general, we expect the retrieval to generate pseudo text features that align with images, and optimize to make the pseudo text feature contain more discriminative semantic information (indicated in parenthesis).}
    \label{fig:template}
    \vspace{-0.2in}
\end{figure*}

To tackle the above mis-alignment issue when pre-training with only image samples, we propose a retrieval-then-optimization method. It consists of two stages: a retrieval-augmented approach to construct initial pseudo features from synthesized text (Section~\ref{sec:retieval}), and a contrastive latent optimization procedure to better align pseudo text features with its paired image (Section~\ref{sec:latentop}). The core idea of our retrieval-then-optimization approach is illustrated in Figure \ref{fig:template}, which can lead to better pseudo text features as illustrated in Figure \ref{fig:similarity}.

\subsection{Pseudo Text-Feature Synthesis: A Retrieval-Augmented Approach}\label{sec:retieval}

In the first stage, we consider a retrieval-augmented approach. It proceeds in four steps, where we take MS-COCO as an example and elaborate the process as follows.

\textbf{Step 1: Vocabulary Construction.}
We construct our vocabulary based on the vocabulary of CLIP, whose size is 49,408. We first filter out non-text items such as emoji and punctuation, then construct several smaller vocabularies in different categories, including: $(i)$ {\it Nouns} that contains names of different places, occupations, creatures, physical objects and abstract concepts such as \texttt{life}. These nouns may serve as either subject or object in a caption; $(ii)$ {\it Verbs} that represent the relations, such as doing, coming, cleaning; $(iii)$ {\it Numerals \& Quantifiers} such as one, two, some, many; $(iv)$ {\it Adjectives} such as red, green, large;
Note that if the pre-training dataset is intentionally collected, one can apply further filtering or construct extra vocabulary for the customized dataset because of prior knowledge. For example, for a customized dataset containing only human faces, we can filter out many nouns and only leave those facial features. We can also construct an extra vocabulary consists of phrases such as wearing glasses, wearing earrings, etc.

\textbf{Step 2: Image-to-Word Retrieval.}
For each word in the constructed vocabularies, we extract its text feature using pre-trained CLIP text encoder. 
We denote the word features as $\{\wb_{ij}\}$, where $\wb_{ij}$ indicates that it is the feature of the $j^{th}$ word from the $i^{th}$ vocabulary. 
Given an image $\xb$, we extract its visual feature using the CLIP image encoder, and compute its cosine similarity with every $\wb_{ij}$. For every vocabulary, the top-$K$ that are most similar to the query image are retrieved.

\textbf{Step 3: Prompt Generation.} Inspired by the success of language prompt in~\cite{radford2021learning}, we build a set of templates in a domain. Our language prompts (\ie fake captions/sentences) are constructed by compositing the retrieved words with the templates. Specifically, the prompt template for MS-COCO is "\texttt{\{Numeral/Quantifier\} \{Adjective\} \{Noun\} \{Verb\}  \{Adjective\} \{Noun\}}". The corresponding words are fitted into the template to generate fake captions, resulting in $K^6$ fake captions.

\textbf{Step 4: Image-to-Prompt Retrieval.}
 The above generated captions are then fed into the CLIP text encoder to obtain corresponding text features, which are used for image-to-prompt retrieval. Only text features that endow high cosine similarity with the given image feature will be used as the corresponding text features for the image, which are guaranteed to lie inside of $\mathcal{T}$ as they are the outputs of the text encoder. 
 
\textbf{Remark.}
The number of fake captions can be very large when $K$ is large or the number of prompt templates is large. Since all the fake captions have to be fed into CLIP text encoder to obtain their text features, more feed-forwards leads to more processing time.
To accelerate, one can implement Step 3 and Step 4 in an iterative manner. For example, we first use a simpler template "\texttt{\{Noun\} \{Verb\} \{Noun\}}", which leads to $K^3$ relations. Then, we selected $K$ out of them by querying with the given image. The selected words are denoted as $\{(\text{Noun}_i, \text{Verb}_i, \text{Noun}_i)\}_{i=1}^K$, and inserted into "\texttt{\{Numeral/Quantifier\} \{Adjective\} $\text{Noun}_i$ $\text{Verb}_i$  \{Adjective\} $\text{Noun}_i$}". We implement the same steps again, which leads to $K^4$ fake captions. There are $(K + 1)K^3$ fake captions generated in total, which is much smaller than $K^6$ for any $K\geq 2$.

The retrieval-augmented process can be conducted once offline.
Compared to the 70K hours spent in constructing MS-COCO dataset~\cite{lin2014microsoft}, the proposed method only takes 5 hours on a single Nvidia Tesla V100 GPU. In practice, we notice that training on only template-based text features may result in over-fitting. Hence, we also apply Gaussian perturbation~\cite{zhou2021lafite} as data augmentation. As shown in the experiments, it prevents over-fitting, thus improves the performance of our pre-trained models.


\subsection{Pseudo Text-Feature Refinement: Contrastive Latent Optimization}\label{sec:latentop}
Our second stage considers the generated pseudo text features in the first stage as initialization, and aims to optimize in the CLIP multi-modal feature space so that they can align better with their paired images.
Following~\cite{zhou2021lafite}, we use the same StyleGAN2-based network architecture~\cite{karras2020analyzing}. The generator takes both text feature and random noise as latent inputs. Let $\{\hb_{ij}\}$ be the corresponding pseudo text features for real image $\xb_i$. 
At each training iteration, a mini-batch of real images are sampled, denoted as $\{\xb_i\}_{i=1}^n$ with $n$ being the mini-batch size.
For each image $\xb_i$, we randomly sample one feature from $\{\hb_{ij}\}$, denoted as $\hb_i^0$. Then $\{\hb_i^0\}_{i=1}^n$ will be refined with \eqref{eq:sampling}, which essentially optimizes a contrastive loss evaluated using $\{(\hb_i^t, \xb_i)\}_{i=1}^n$.
\begin{align}\label{eq:sampling}
    \hb_i ^{t+1} &= \hb_i ^{t} + \lambda \nabla_{\hb_i ^{t}} \sum_{i=1}^n \log c_{ii}^t, ~~~~ t\leq T-1\nonumber 
    \\c_{ii}^t &= \dfrac{\exp(\text{Sim}(f_{\text{img}}(\xb_i), \hb_i ^t )/\tau)}{\sum_{k=1}^n \exp(\text{Sim}(f_{\text{img}}(\xb_k), \hb_i ^t) /\tau)},
\end{align}


where $T \geq 0$ denotes the number of latent optimization updates, $\lambda>0$ is the step size,  and ``$\text{Sim}$'' denotes cosine similarity. Intuitively, \eqref{eq:sampling} drives a text feature $\hb_i^t$ to be more similar with the corresponding ground-truth image feature $f_{\text{img}}(\xb_i)$, while imposing dissimilarities with other image features $\{f_{\text{img}}(\xb_j)\}_{j \neq i}$.  

\paragraph{Theoretical Justification.}
We explain the underlying theoretical principle of the proposed contrastive latent optimization (CLO) for better image-text alignment. \cite{zhou2021lafite} demonstrates that model performance can be significantly improved with contrastive loss. Specifically, we are given a mini-batch of pairs $\{(\xb_i^\prime, \hb_i)\}_{i=1}^n$, where $\xb_i^\prime$ and $\hb_i$ respectively denote the generated image and input feature. The loss for the generator $G_{\thetab}$ is defined as a combination of the standard adversarial loss and an additional contrastive loss:
\begin{align}\label{eq:contrastive_loss}
    \mathcal{L} &= -\sum_{i=1}^n \mathcal{L}_i\nonumber \\
    \mathcal{L}_i &= \log \dfrac{\exp (\text{Sim}(f_{\text{img}}(\xb_i^\prime), \hb_i)/\tau)}{\sum_{j=1}^n \exp(\text{Sim}(f_{\text{img}}(\xb_j^\prime), \hb_i)/\tau)}, \nonumber \\
    \xb^\prime &= G_{\thetab}(\hb_i, \epsilonb_i)
\end{align}
where $\tau>0$ is a hyper-parameter, and $\{\epsilonb_i\}_{i=1}^n$ denote random noises.
Intuitively, \eqref{eq:contrastive_loss} enforces the generated image $\xb_i^\prime$ to have high cosine similarities with the corresponding input text features $\hb_i$ in the multi-modal feature space of CLIP, while having low similarities with other features $\{\hb_j\}_{j\neq i}$. Essentially, the contrastive loss provides knowledge of image-text correspondence for the generator, making it better for text-to-image generation.
However, we find the gradient norm, $\Vert \nabla_{\thetab} \mathcal{L}\Vert_2$, could be too small to provide enough information to the generator under certain circumstances. 
Specifically, we can prove Theorem \ref{thm:gradient_bound} below:

\begin{theorem}\label{thm:gradient_bound}
    Let $\{\xb_j^\prime\}_{i=1}^n$ be a mini-batch of generated images, $\{\hb_i\}_{i=1}^n$ be the corresponding text features fed into the generator $G_{\thetab}$. For the contrastive loss $\mathcal{L}$ in \eqref{eq:contrastive_loss}, we have
    $\Vert \nabla_{\thetab} \mathcal{L}\Vert_2 \leq na + n^2a\sigma
    $, where 
    $a\geq 0$ is a constant depending on the CLIP image encoder and our generator, and $\sigma \geq 0$ denotes the standard deviation of $\{c_{ji}\}$ with $c_{ji} = \exp(\text{Sim}(f_{\text{img}}(\xb_j^\prime), \hb_i)/\tau)/\sum_{k=1}^n \exp(\text{Sim}(f_{\text{img}}(\xb_k^\prime), \hb_i)/\tau)$.
\end{theorem}
Theorem~\ref{thm:gradient_bound} indicates that the gradient norm is bounded by a value which increases along with $\sigma$. We can easily find that the smallest $\sigma=0$ is obtained when $\text{Sim}(f_{\text{img}}(\xb_j^\prime), \hb_i) = b$ for all $i, j$ with $b$  being a constant. 
Intuitively, small $\sigma$ means that an input text feature is indistinguishable for the generated images. Thus, the contrastive loss can not provide useful information because the gradient norm is too small. This could happen in many scenarios including: $(i)$ in the fully-supervised setting, text captions are not precise and informative enough; $(ii)$ in the unsupervised setting, the generated text features could be inappropriate as there is no monitoring and correction; $(iii)$ the generated images $\{\xb^\prime_j\}_{j=1}^n$ are vague and inaccurate, which is common at the early stage of training.
With our constrastive latent optimization technique, the generated pseudo text features are updated so that they are more distinctive, which normally could lead to larger variance.
Note that we do not update samples by the gradient $\nabla_{\hb_i} \sigma^2$, because we want to make sure that the image-text correspondence is correctly maintained; whereas simply updating w.r.t $\sigma^2$ may damage this correspondence.



\begin{table*}[ht!]
    \centering
    \caption{Fully-supervised text-to-image generation task on different datasets. \setlength{\fboxsep}{0pt}\colorbox{Gray}{\strut Gray cells} are evaluated using publicly available pre-trained models provided by the authors.}
    \label{tab:results_sup}
    \scalebox{0.75}{
    \begin{tabular}{l|cccccccccc}
    \toprule
        & \multicolumn{4}{c}{MS-COCO} &\multicolumn{2}{c}{CUB} & \multicolumn{2}{c}{LN-COCO} & \multicolumn{2}{c}{MM-CelebA-HQ}\\
         Model &   IS $(\uparrow)$  & FID $(\downarrow)$  &  SOA-C $(\uparrow)$ &  SOA-I ($\uparrow$) &   IS $(\uparrow)$  &  FID $(\downarrow)$  &  IS $(\uparrow)$  &  FID $(\downarrow)$  &  IS $(\uparrow)$ & FID $(\downarrow)$ \\ 
        \midrule
         AttnGAN~\cite{xu2018attngan}& $23.61$ & $33.10$ & $25.88$ & $39.01$ & $4.36$ & $23.98$ & $20.80$ & $51.80$ & - & $125.98$\\
         Obj-GAN~\cite{li2019object} & $24.09$ & $36.52$ & $27.14$ & $41.24$ &-&-&-&-\\
         DM-GAN~\cite{zhu2019dm} & $32.32$ & $27.34$ & $33.44$ & $48.03$ & $4.75$ & $16.09$ & -& -& - &$131.05$ \\
         OP-GAN~\cite{hinz2020semantic} & $27.88$ & $24.70$ & $35.85$ & $50.47$ & - & - & - & - & - & -\\
         DF-GAN~\cite{tao2021dfgan} & - & $21.42$ & - & -  & $5.10$ & $14.81$ & - & - & - & $137.60$ \\
         N\"UWA~\cite{wu2021n} & $27.20$ & $12.90$ & & & & & & & \\
         XMC-GAN~\cite{zhang2021crossmodal}  & $30.45$ & $9.33$ & $50.94$ & $71.33$ & - & - & $28.37$ & $14.12$ & - & - \\
         \shortnamecvpr{}~\cite{zhou2021lafite}  &  $32.34$ & $8.12$ & $61.09$ & $74.78$ &  $5.97$ & $10.48$ & $26.32$ & $11.78$ & $2.93$ & $12.54$\\
         Make-A-Scene~\cite{gafni2022make} & - & $7.55$ & - &- & - &- & - & - & - & -\\
         LDM~\cite{rombach2021high}& \cellcolor{Gray}{$31.42$}& \cellcolor{Gray}{$5.97$} & - &- & - &- & - & - & - & -\\      
          Parti~\cite{yu2022scaling}& - & $3.22$ & - &- & - &- & - & - & - & -\\
          Re-Imagen~\cite{chen2022re}& - & $5.25$ & - &- & - &- & - & - & - & -\\
         \midrule
        \rowcolor{unitednationsblue!20} 
         \lafitegan{}  &$\mathbf{34.16}$ & $\mathbf{6.78}$ & $\mathbf{61.30}$ & $\mathbf{74.84}$ & $\mathbf{6.06}$ & $\mathbf{8.59}$ & $\mathbf{28.71}$ & $\mathbf{8.85}$ & $\mathbf{3.17}$ & $\mathbf{8.22}$\\
        \rowcolor{unitednationsblue!20} 
         \lafiteldm{}  & $\mathbf{31.74}$ & $\mathbf{4.28}$ & - &- & - &- & - & - & - & -\\         
         \bottomrule
    \end{tabular}
    }
\end{table*}

\subsection{\shortname{} Model Instantiation}\label{sec:lafite2_models}
The proposed pseudo text feature synthesis techniques in this paper are general, and can be plugged into existing T2I methods to improve their performance. We demonstrate it with two popular model families.

\paragraph{\lafitegan{}.} StyleGAN and its followups~\cite{karras2019style, karras2020analyzing, karras2021alias} are a family of most powerful GAN models for unsupervised or label-conditional image generation, which was adjusted for T2I task in \shortnamecvpr{}~\cite{zhou2021lafite}. For fair comparison, we adopt the same network architecture with \shortnamecvpr{}~\cite{zhou2021lafite}, and replace the  pseudo text features with the ones generated by the proposed techniques in this paper. For a given dataset, we first construct pseudo captions for every image with the proposed method. CLIP text encoder is then used to obtain text features of the constructed captions. The text features will be updated by \eqref{eq:sampling} and injected into the Style Space~\cite{wu2021stylespace} of StyleGAN2. Since the Style Space is well-disentangled~\cite{wu2021stylespace}, injecting text features into Style Space will effectively force the generated images to be text-aligned.

\paragraph{\lafiteldm{}.} Diffusion-based methods are arguably the most effective model family for the T2I tasks in the recent development, among which LDM~\cite{rombach2021high} and its successor Stable Diffusion are the most powerful models publicly available for the research community. Therefore, we apply the proposed retrieval-augmented technique to LDM, whose training objective solves the the denosing problem on latent representations $\zb$ of the image $\xb$: 
\begin{align}\label{eq:ldm_loss}
\mathcal{L}_{\text{LDM}} = \mathbb{E}_{\zb, \epsilonb \sim  \mathcal{N}(\mathbf{0}, \mathbf{I}), t} \big[ \|  \epsilonb -  f_{\text{denoise}} (\zb_t, t, \yb) \|^2_2 \big],
\end{align}
where $t$ is uniformly sampled from time steps $\{1, \cdots, T\}$, $\zb_t$ is the step-$t$ noisy variant of input $\zb$, $\yb$ is the caption condition, and $ f_{\text{denoise}} (*, t, \yb)$ is the $(t, \yb)$-conditioned denoising autoencoder implemented via UNet~\cite{ronneberger2015u}.  In the original LDM, BERT~\cite{devlin2019bert} is utilized to obtain a sequence of text embeddings for each caption, $f_{\text{bert}}(\yb)$, which is fed into~\eqref{eq:ldm_loss} to replace $\yb$. Time $t$ is first mapped to time embedding $\phi(t)$, then injected into the UNet.
Different from the original LDM, our \lafiteldm{} model introduces one extra input for the denoising process: the CLIP text feature $f_{\text{text}}(\yb)$. The CLIP text features are first projected with a linear layer $\mathbf{P}$ to align the dimension, then added to the time embedding via $\phi'(t) = \mathbf{P}f_{\text{text}}(\yb) + \phi(t)$. $\phi'(t)$ is used to replace the original time embedding $\phi(t)$. During the unsupervised pre-training stage, we only use the synthetic captions produced by the proposed retrieval-augmented approach in Section~\ref{sec:retieval}. In the fine-tuning and evaluation stage with ground-truth image-text pairs, the ground-truth captions are used. 


\vspace{0mm}
\section{Experiments}
\vspace{-2mm}
In this section, we examine the proposed approach to answer two research questions. 
\textbf{\texttt{Q1}}: When pre-training and  fine-tuning in a single dataset,  how does our method benefit fully- and semi-supervised learning a particular domain? (Section~\ref{sec:exp_in_domain})
\textbf{\texttt{Q2}}: When pre-training on a general corpus then adapting to downstream datasets, does \shortname{} help zero-shot and few-shot task-level transfer? (Section~\ref{sec:exp_cross_domain})

\begin{table}[ht!]
    \centering
    \caption{In-domain pre-training: semi-supervised text-to-image generation on different datasets.}
    \label{tab:semi-main}
    \scalebox{0.5}{
    \begin{tabular}{lcccccccccc}
    \toprule
         &  \multicolumn{2}{c}{0-shot (language-free)} &  \multicolumn{2}{c}{10-shot} & \multicolumn{2}{c}{20-shot} & \multicolumn{2}{c}{50-shot} & \multicolumn{2}{c}{100-shot} \\
         Model & IS $(\uparrow)$ & FID $(\downarrow)$ & IS $(\uparrow)$ & FID $(\downarrow)$ & IS $(\uparrow)$ & FID $(\downarrow)$ & IS $(\uparrow)$ & FID $(\downarrow)$ & IS $(\uparrow)$ & FID $(\downarrow)$ \\
        \midrule
        & \multicolumn{8}{c}{MS-COCO}\\
         SEMI & \multirow{2}{*}{27.20} & \multirow{2}{*}{18.04}& $27.26$ & $18.00$ & $27.28$ & $17.98$ & $27.20$ & $18.16$ & $27.28$ & $18.00$ \\
         \shortnamecvpr{}~\cite{zhou2021lafite} & & & $27.47$ & $18.02$ & $27.70$ & $17.51$ & $28.18$ & $17.13$ & $27.58$ & $16.82$ \\
         \rowcolor{unitednationsblue!20} 
         \lafitegan{}  & $\mathbf{32.16}$ & $\mathbf{10.26}$ & $\mathbf{32.10}$ & $\mathbf{10.20}$ & $\mathbf{32.20}$ & $\mathbf{10.16}$ & $\mathbf{31.94}$ & $\mathbf{10.11}$ & $\mathbf{32.25}$ & $\mathbf{9.84}$ \\
         \midrule
         & \multicolumn{8}{c}{CUB}\\
         SEMI & \multirow{2}{*}{4.32} & \multirow{2}{*}{27.53} & $4.31$ & $26.05$ & $4.33$ & $25.61$ & $4.37$ & $25.39$ & $4.45$ &$24.20$ \\
         \shortnamecvpr{}~\cite{zhou2021lafite} & & & $4.27$ & $25.08$ & $4.48$ & $24.93$ & $4.49$ & $22.51$ & $4.96$ & $22.01$\\
         \rowcolor{unitednationsblue!20} 
         \lafitegan{}  & $\mathbf{4.93}$ & $\mathbf{16.87}$ & $\mathbf{5.14}$ & $\mathbf{16.26}$ & $\mathbf{5.20}$ & $\mathbf{16.00}$ & $\mathbf{5.24}$ & $\mathbf{14.16}$ & $\mathbf{5.13}$ & $\mathbf{14.11}$\\
         \midrule
         & \multicolumn{8}{c}{LN-COCO}\\
         SEMI & \multirow{2}{*}{18.49} & \multirow{2}{*}{39.85} & $21.09$ & $34.96$ & $21.08$ & $34.84$ & $21.10$ & $34.65$ & $21.25$ & $34.20$ \\
         \shortnamecvpr{}~\cite{zhou2021lafite} & & & $22.50$ & $28.57$ & $22.31$ & $27.90$ & $23.56$ & $23.89$ & $24.50$ & $22.17$ \\
         \rowcolor{unitednationsblue!20} 
          \lafitegan{}  & $\mathbf{23.18}$ & $\mathbf{25.51}$& $\mathbf{23.95}$ & $\mathbf{23.65}$ & $\mathbf{24.04}$ & $\mathbf{22.74}$ & $\mathbf{24.37}$ & $\mathbf{19.63}$ & $\mathbf{25.12}$ & $\mathbf{18.49}$\\
         \midrule
         & \multicolumn{8}{c}{MM-CelebA-HQ}\\
         SEMI & \multirow{2}{*}{3.10} & \multirow{2}{*}{15.74} & $3.03$ & $14.82$ & $3.06$ & $14.79$ & $3.04$ & $14.60$ & $3.05$ & $14.64$ \\
         \shortnamecvpr{}~\cite{zhou2021lafite} & & & $3.00$ & $14.80$ & $2.99$ & $14.19$ & $3.01$ & $14.02$ & $3.00$ & $13.55$\\
         \rowcolor{unitednationsblue!20} 
         \lafitegan{}  & $\mathbf{3.37}$ & $\mathbf{11.66}$ & $\mathbf{3.36}$ & $\mathbf{10.97}$ & $\mathbf{3.32}$ & $\mathbf{10.87}$ & $\mathbf{3.24}$ & $\mathbf{10.54}$ & $\mathbf{3.33}$ & $\mathbf{9.94}$\\
         \bottomrule
    \end{tabular}
    \vspace{-0.3in}
    }
\end{table}
\paragraph{Settings \& Evaluation Metrics.} All the experiments are conducted on 4 Nvidia Tesla V100 GPUs, implemented with Pytorch~\cite{paszke2019pytorch}. We follow the same network architecture in \shortnamecvpr{}~\cite{zhou2021lafite} and LDM~\cite{rombach2021high} for fair comparisons. Standard datasets are considered for downstream tasks, including MS-COCO~\cite{lin2014microsoft}, CUB~\cite{WahCUB_200_2011}, MM-CelebA-HQ~\cite{xia2021tedigan}, LN-COCO~\cite{pont2020connecting}. The data statistics are shown in Appendix. To quantitatively measure the image generation quality, we report Fr\'echet Inception Distance (FID) \cite{heusel2017gans} and Inception Score (IS) \cite{salimans2016improved}, which are computed with 30K generated images whose text inputs are randomly sampled from the validation set of MS-COCO, CUB and LN-COCO datasets, 6K images are generated for MM-CelebA-HQ following \cite{xia2021tedigan,zhou2021lafite}. We also report Semantic Object Accuracy (SOA) on MS-COCO dataset following previous works~\cite{pont2020connecting,zhang2021crossmodal,zhou2021lafite}, which evaluates whether the generated images contain the desired objects. Some generated examples are provided in the appendix.
All the code and pre-trained models will be made publicly available.

\subsection{Unsupervised Pre-training}\label{sec:exp_in_domain}
In this experiment, our model is first pre-trained with pseudo image-text pairs generated with the proposed method, then fine-tuned with ground-truth image-text pairs from the same dataset.

We first study the fully-supervised setting in which all the ground-truth image-text pairs are provided, and compare the proposed method with current SoTA across different datasets. 
$(i)$ 
For \lafitegan{}, the model is first pre-trained on training images from downstream dataset paired with pseudo text features (which is denoted as {\it in-domain} pre-training), then fine-tuned with ground-truth text features. 
$(ii)$ 
For \lafiteldm{}, the model is first pre-trained on a set of unlabelled images, which has no overlap with downstream dataset (which is denoted as {\it near-domain} pre-training), then fine-tuned with ground-truth image-text pairs from downstream dataset. Our \lafiteldm{} is initialized with the publicly available checkpoint from ~\cite{rombach2021high} with 1.4 billions of parameters, and we only update the UNet part which contains 0.87 billions of parameters, and freeze the transformer part which process the text embeddings before feeding them into UNet.
We compare our methods with different fully-supervised methods including AttnGAN~\cite{xu2018attngan}, Obj-GAN~\cite{li2019object}, DM-GAN~\cite{zhu2019dm}, OP-GAN~\cite{hinz2020semantic}, DF-GAN~\cite{tao2021dfgan}, XMC-GAN~\cite{zhang2021crossmodal}, \shortnamecvpr{}~\cite{zhou2021lafite} and Make-A-Scene~\cite{gafni2022make}. 
Note \cite{rombach2021high} only provided zero-shot text-to-image generation results on MS-COCO dataset, to report fully-supervised performance of LDM, we fine-tuned the checkpoint provided by the authors on MS-COCO dataset. The main results are reported in Table \ref{tab:results_sup}, which shows that our method outperforms other methods in all metrics. For \lafiteldm{}, the checkpoint at fine-tuning step 10K is reported. We found it provides slightly better FID results than training longer, \eg 100K steps.


We then consider more flexible settings: semi-supervised text-to-image generation, where all the downstream training images are provided while only few of them have corresponding captions. It is more practical because one can choose the number of images to be captioned based on their available resources and budget. Specifically,
10, 20, 50, 100 image-text pairs are provided to fine-tune the unsupervised pre-trained model. Freeze-D ~\cite{mo2020freeze} is used during fine-tuning for \lafitegan{}.
For quantitative evaluation, we choose two different approaches as baseline methods: $(i)$ \shortnamecvpr{}: pre-training a model with pseudo image-text pairs generated by \shortnamecvpr{}, then fine-tune it with the provided pairs; 
$(ii)$ SEMI: directly training the model from scratch, with a mixture of ground-truth and pseudo pairs which are generated by \shortnamecvpr{}. When no ground-truth pair is provided, these two baselines become the same.
The results are shown in Table \ref{tab:semi-main}, our method significantly outperforms the baselines in all few-shot cases.

\begin{table*}[t!]
    \centering
    \caption{Zero-shot text-to-image generation. LDM-G denotes latent diffusion model with classifier-free guidance. \setlength{\fboxsep}{0pt}\colorbox{Gray}{\strut Gray cells} are evaluated using publicly available pre-trained models provided by the authors. For fair comparison, the batch size is set to be 1 for all the models when we compute the inference time, which are averaged over 1000 generations.
    250 steps of DDIM~\cite{song2020denoising} sampling is used along with classifier-free guidance~\cite{ho2021classifier} in LDM, which follows ~\cite{rombach2021high} for fair comparison.
    $^{\dagger}$4.6B parameter frozen text encoder (T5-XXL) is used. }
    \label{tab:zero-shot}
    \scalebox{0.75}{
    \begin{tabular}{lllcccc}
    \toprule
         \multirow{2}{*}{Model} &  \#Trainable & Inference Time & \multicolumn{4}{c}{FID  $(\downarrow)$ }\\
          &  Parameters  &  (ms) & MS-COCO & CUB & LN-COCO  & MM-CelebA-HQ \\
        \midrule
        Parti~\cite{yu2022scaling} & 20B & & $7.23$ & - & - & -\\
        DALL-E~\cite{ramesh2021zero} & 12B & & $27.50$ & - & - & -\\
        GLIDE~\cite{nichol2021glide} & 6B & $1500$ & $12.24$ & - & - & - \\
        DALL-E 2~\cite{ramesh2022hierarchical} & 5.5B & & $10.39$ & - & - & - \\
        Make-A-Scene~\cite{gafni2022make} & 4B & & $11.84$ & - & - & - \\
        CogView~\cite{ding2021cogview} & 4B & & $27.10$ & - & - & - \\
        Re-Imagen~\cite{chen2022re} & & & $6.88$ & - & - & - \\
        Imagen~\cite{saharia2022photorealistic} & 3B$^{\dagger}$ & & $7.27$ & - & - & - \\
        LDM~\cite{rombach2021high} & 1.45B & \cellcolor{Gray}{$2836$ } & $23.31$ & -  & - & -\\
        LDM-G~\cite{rombach2021high} & 1.45B & \cellcolor{Gray}{$20833$ } & $12.63$ /  \colorbox{Gray}{$11.73$} & - & - & - \\
        KNN-Diffusion~\cite{ashual2022knn} & 470M & $800$ & $16.66$ & - & - & - \\
        \shortnamecvpr{}~\cite{zhou2021lafite} & 75M &  \cellcolor{Gray}{43 } & $26.94$ &  \cellcolor{Gray}{$68.53$} & \cellcolor{Gray}{$36.91$} & \cellcolor{Gray}{$71.69$} \\
         \midrule
         \rowcolor{unitednationsblue!20} 
         \lafitegan{}  & 75M & 43  & $13.42$ & $32.19$ & $25.44$ & $39.46$\\
         \rowcolor{unitednationsblue!20} 
         \lafitegan{} (ND) & 75M & 43 & $10.71$ & $\mathbf{32.04}$ & $23.32$ & $\mathbf{34.93}$ \\
         \lafiteldm{} (ND) & 1.45B & $20833$ & $8.42$ &  - & - & - \\         
         \bottomrule
    \end{tabular}
    }
\end{table*}

\begin{figure*}
    \centering
    \includegraphics[width=0.85\linewidth]{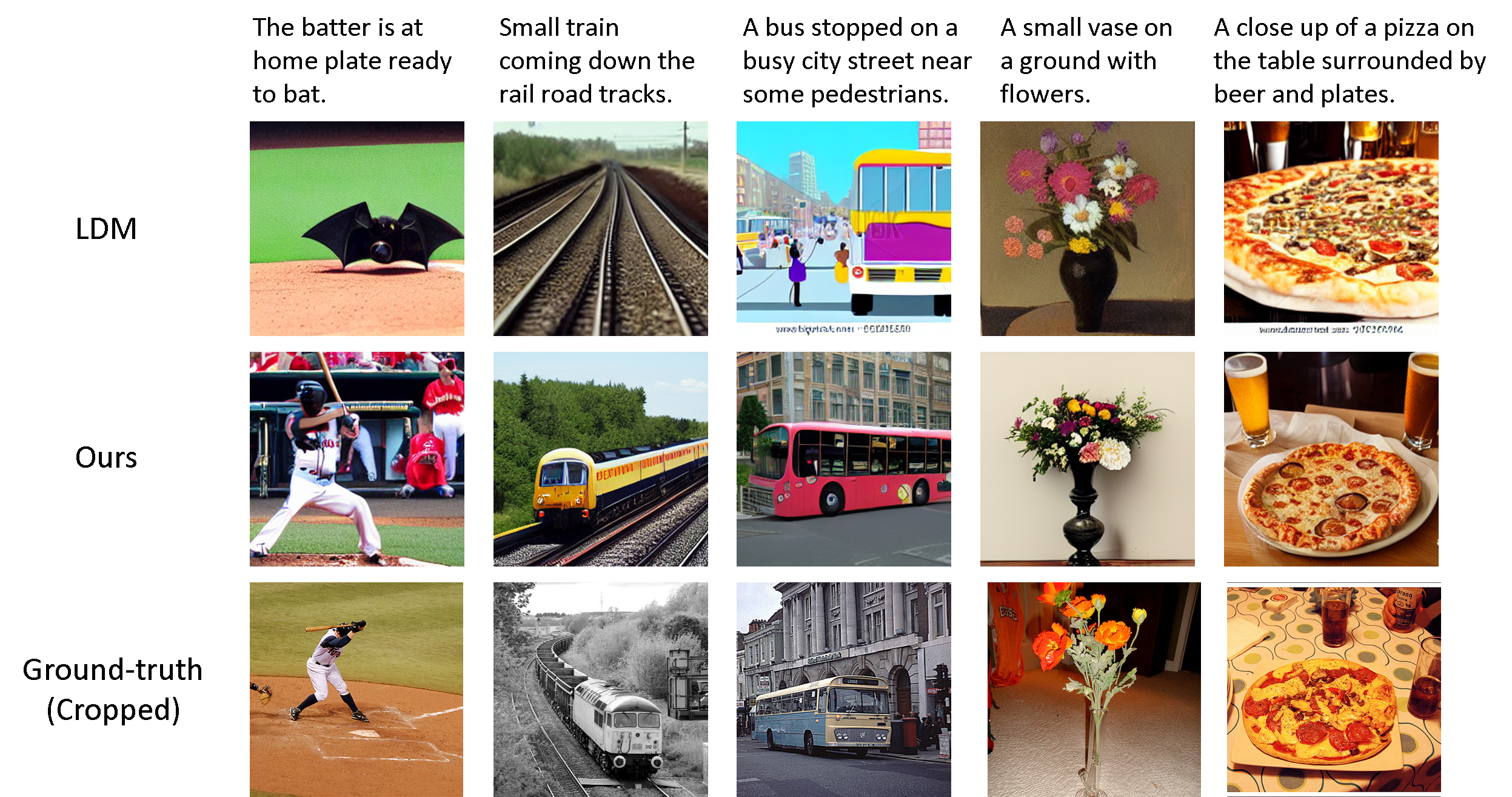}
    \caption{Generated examples with captions from MS-COCO validation set. With the proposed method, our LDM leads to better zero-shot generation. The generated images have better quality and desired styles. }
    \label{fig:ldm_comparison}
\end{figure*}
\subsection{Zero-shot and Few-shot Task Transfer}\label{sec:exp_cross_domain}
For task-level transfer, there is no image overlapping between pre-training datasets and downstream datasets.
%
Depending on the pre-training data, we prepare two variants of the proposed method, denoted as \shortname{} and \shortname{} (ND), where ND stands for near-domain, meaning the pre-training datasets are selected to be related to the downstream dataset. 
For example, when the downstream dataset is the CUB dataset that contains image-text pairs of various birds, we can use Birdsnap dataset~\cite{Berg_2014_CVPR} as the pre-training dataset, which is an image-only dataset containing 49,829 bird images.
More details can be found in Appendix. 

\begin{table*}[ht!]
    \centering
    \caption{Few-shot text-to-image generation on four datasets. For fair comparisons, we also pre-train \shortnamecvpr{}~\cite{zhou2021lafite} on the same datasets with \shortname{}, denoted as \shortnamecvpr{}$^*$ and \shortnamecvpr{}-ND$^*$
    }
    \label{tab:few-shot}
    \scalebox{0.8}{
    \begin{tabular}{lcccccccc}
    \toprule
         &  \multicolumn{2}{c}{10-shot} & \multicolumn{2}{c}{20-shot} & \multicolumn{2}{c}{50-shot} & \multicolumn{2}{c}{100-shot} \\
         Model & IS $(\uparrow)$ & FID $(\downarrow)$ & IS $(\uparrow)$ & FID $(\downarrow)$ & IS $(\uparrow)$ & FID $(\downarrow)$ & IS $(\uparrow)$ & FID $(\downarrow)$\\
        \midrule
        & \multicolumn{8}{c}{MS-COCO}\\
        From scratch & $3.29$ & $216.34$ & $3.66$ & $212.39$ & $4.43$ & $188.45$ & $4.62$ & $182.42$\\
        \shortnamecvpr{}~\cite{zhou2021lafite} & $25.18$ & $24.39$ & $25.73$ & $22.03$ & $25.96$ & $21.55$ & $26.43$ & $18.77$\\
        \shortnamecvpr{}$^*$ & $19.37$ & $36.52$ & $20.09$ & $34.94$ & $20.95$ & $34.31$ & $20.94$ & $31.94$\\
        \rowcolor{unitednationsblue!20} 
          \lafitegan{} & $27.20$ & $13.37$ & $27.41$ & $13.38$ & $27.36$ & $13.32$& $27.58$ & $13.19$\\           
        \shortnamecvpr{} (ND)$^*$ &  $27.72$ & $19.72$ & $27.65$ & $19.06$ & $28.27$ & $18.43$ & $28.43$ & $17.91$\\
         \rowcolor{unitednationsblue!20} 
        \lafitegan{} (ND)  & $\mathbf{31.46}$ & $\mathbf{10.67}$ & $\mathbf{31.80}$ & $\mathbf{10.63}$ & $\mathbf{31.72}$ & $\mathbf{10.57}$ & $\mathbf{32.30}$ & $\mathbf{10.31}$\\
        LDM  & -  & - &  - & - &  25.11 &  10.45 & 24.84 &  10.30 \\    
        \rowcolor{unitednationsblue!20} 
        \lafiteldm{} (ND)  & -  & - &  - & - &  $\mathbf{30.36}$  & $\mathbf{6.26}$ &  $\mathbf{30.09}$ & $\mathbf{6.05}$ \\        
         \midrule
         & \multicolumn{8}{c}{CUB}\\
        From scratch & $5.16$ & $127.88$ & $5.59$ & $125.49$ & $5.54$ & $116.81$ & $7.16$ & $115.05$\\
        \shortnamecvpr{}~\cite{zhou2021lafite} & $3.90$ & $55.41$ & $4.36$ & $53.67$ & $3.93$ & $52.39$ & $4.52$ & $46.27$\\
        \shortnamecvpr{}$^*$ & $3.87$ & $30.10$ & $3.93$ & $28.79$ & $4.08$ & $25.13$ & $4.25$ & $23.26$\\
        \rowcolor{unitednationsblue!20} 
         \lafitegan{} & $5.77$ & $26.86$ & $5.68$ & $26.79$  & $5.51$ & $23.53$ & $5.39$ & $20.48$ \\
        \shortnamecvpr{}-ND$^*$ & $5.08$ & $26.87$ & $5.17$ & $26.08$ & $5.11$ & $24.79$ & $4.78$ & $22.69$\\ 
         \rowcolor{unitednationsblue!20} 
         \lafitegan{} (ND)  & $6.56$ & $28.07$ & $6.50$ & $26.87$ & $6.40$ & $20.84$ & $5.64$ & $18.44$\\
         \midrule
         & \multicolumn{8}{c}{LN-COCO}\\
         From scratch & $2.49$ & $269.96$ & $3.24$ & $260.47$ & $2.98$ & $254.08$ & $3.58$ & $223.59$\\
        \shortnamecvpr{}~\cite{zhou2021lafite} & $18.84$ & $34.09$ & $18.69$ & $32.94$ & $18.81$ & $31.13$ & $19.48$ & $28.81$\\
        \shortnamecvpr{}$^*$ & $15.40$ & $48.02$ & $16.03$ & $42.84$ & $15.91$ & $42.41$ & $16.23$ & $40.51$\\
        \rowcolor{unitednationsblue!20} 
        \lafitegan{} & $21.20$ & $25.70$ & $21.23$ & $25.57$ & $21.47$ & $24.89$ & $21.46$ & $24.24$\\
        \shortnamecvpr{}-ND$^*$ & $21.74$ & $31.52$ & $22.51$ & $28.71$ & $22.53$ & $25.82$ & $23.45$ & $25.28$\\
         \rowcolor{unitednationsblue!20} 
        \lafitegan{} (ND) & $\mathbf{23.79}$ & $\mathbf{22.78}$ & $\mathbf{23.54}$ & $\mathbf{21.43}$ & $\mathbf{24.36}$ & $\mathbf{19.03}$ & $\mathbf{24.81}$ & $\mathbf{17.83}$\\
         \midrule
         & \multicolumn{8}{c}{MM-CelebA-HQ}\\
         From scratch & $2.43$ & $170.64$ & $2.48$ & $143.09$ & $2.48$ & $99.79$ & $2.40$ & $87.91$ \\
         \shortnamecvpr{}~\cite{zhou2021lafite} & $3.40$ & $56.54$ & $3.66$ & $55.88$ & $3.54$ & $48.72$ & $3.34$ & $43.33$\\
         \shortnamecvpr{}$^*$ & $3.14$ & $35.80$ & $3.06$ & $33.04$ & $3.14$ & $30.92$ & $3.11$ & $31.81$\\
        \rowcolor{unitednationsblue!20} 
         \lafitegan{} & $3.15$ & $34.86$ & $3.03$ & $29.06$ & $3.26$ & $25.79$ & $3.05$ & $23.12$\\
         \shortnamecvpr{} (ND)$^*$ & $3.15$ & $27.59$ & $2.91$ & $25.62$ & $3.04$ & $22.34$ & $3.10$ & $21.33$\\
         \rowcolor{unitednationsblue!20} 
         \lafitegan{} (ND) & $3.26$ & $\mathbf{24.33}$ & $3.14$ & $\mathbf{21.08}$ & $3.10$ & $\mathbf{19.90}$ & $3.20$ & $\mathbf{17.68}$\\
         \bottomrule
    \end{tabular}
    }
\end{table*}

\begin{table*}[ht!]
    \centering
    \caption{The FID and IS values over different iterations, by adapting LDM and \lafiteldm{} in 50- and 100-shot settings.
    }
    \label{tab:few-shot_2}
    \scalebox{0.8}{
    \begin{tabular}{llcccccccc}
    \toprule
        &  \multirow{2}{*}{ Model }  &  \multicolumn{2}{c}{0 iteration (zero-shot)} & \multicolumn{2}{c}{400-iteration} & \multicolumn{2}{c}{800-iteration} & \multicolumn{2}{c}{2000-iteration} \\
        & &  FID $(\downarrow)$  & IS $(\uparrow)$ & FID $(\downarrow)$ & IS $(\uparrow)$ & FID $(\downarrow)$ & IS $(\uparrow)$ & FID $(\downarrow)$  & IS $(\uparrow)$\\
        \midrule
        \multirow{2}{*}{ 50-shot } & 
        LDM~\cite{rombach2021high} & $12.63$ & $23.11$
        & 11.38 & 24.22 & 10.45 & 25.11 & 10.57 & 27.05 \\
         & 
         \cellcolor{unitednationsblue!20}\lafiteldm{} &  \cellcolor{unitednationsblue!20}$8.42$ &  \cellcolor{unitednationsblue!20}$24.29$ & 
          \cellcolor{unitednationsblue!20}$6.27$ &   \cellcolor{unitednationsblue!20}$29.45$ &     
          \cellcolor{unitednationsblue!20}$6.26$ & \cellcolor{unitednationsblue!20}$30.36$ &  
          \cellcolor{unitednationsblue!20}$7.14$ & \cellcolor{unitednationsblue!20}$31.88$    \\
          \midrule
        \multirow{2}{*}{ 100-shot } & 
        LDM~\cite{rombach2021high} &  $12.63$   & $23.11$
        & $11.55$ &	$23.48$&	$10.30$&	$24.84$&	$11.19$&	$26.70$ \\
         & 
         \cellcolor{unitednationsblue!20}\lafiteldm{} & \cellcolor{unitednationsblue!20}$8.42$ & \cellcolor{unitednationsblue!20} $24.29$
         & 
          \cellcolor{unitednationsblue!20}6.23   & \cellcolor{unitednationsblue!20}29.89 & 
          \cellcolor{unitednationsblue!20}6.05 & \cellcolor{unitednationsblue!20}30.09 &
          \cellcolor{unitednationsblue!20}7.47 & \cellcolor{unitednationsblue!20}31.80\\
         \bottomrule
         \vspace{1mm}
    \end{tabular}
    }    
\end{table*}

\paragraph{Zero-shot.}
The zero-shot text-to-image generation results are reported in Table \ref{tab:zero-shot}. Our method achieves comparable results with DALL-E 2, but with much smaller model size. Compared with auto-regressive models and diffusion models that often leads to SoTA performance, the proposed \lafitegan{} enjoys favorable property of much lower inference time.
\lafiteldm{} leads to better performance than \lafitegan{} with larger model size and higher inference time. We compare \lafiteldm{} with original LDM in figure \ref{fig:ldm_comparison}, from which we can see that the improvement is obvious: with the proposed method, we can generate images with better quality and avoid generating undesirable image styles. More specifically, since we would like to generate images similar to samples from MS-COCO, we would like the generated images to be realistic photos instead of cartoon images or oil paintings. With the proposed method, we can enforce the LDM generate desired text-aligned images by utilizing a near-domain image-only dataset. The workload of constructing such a dataset can be neglected compared to the performance gain, as it only requires image samples without captions.

\paragraph{Few-shot.}
Few-shot text-to-image generation is more practical but less explored area. 
As shown in Table \ref{tab:few-shot}, training from scratch is challenging under few-shot setting and can not obtain satisfactory results. On the contrary, fine-tuning pre-trained models can lead to much better results. 
The results are shown in Table \ref{tab:few-shot}, where 10, 20, 50, 100 training image-text pairs are provided for model adaptation. 
We also compare the adaptation process on LDM in Table \ref{tab:few-shot_2}. From the results we can conclude that $(i)$ our pre-training is more general and achieves better few-shot text-to-image generation via fine-tuning; $(ii)$ our pre-trained model is easier to fine-tune and leads to better results than uGAN, \shortnamecvpr{} and LDM.



\subsection{Ablation Studies}
\begin{table*}[ht!]
    \centering
    \caption{Ablation studies. Image generation quality on MS-COCO.}\label{tab:ablation_1}
    \scalebox{0.8}{
    \begin{tabular}{lcccccccccc}
    \toprule
         &  \multicolumn{2}{c}{$0\%$} & \multicolumn{2}{c}{$0.1\%$} & \multicolumn{2}{c}{$1\%$} & \multicolumn{2}{c}{$10\%$} & \multicolumn{2}{c}{$100\%$}\\
         Component & IS $(\uparrow)$ & FID $(\downarrow)$ & IS $(\uparrow)$ & FID $(\downarrow)$ & IS $(\uparrow)$ & FID $(\downarrow)$ & IS $(\uparrow)$ & FID $(\downarrow)$ & IS $(\uparrow)$ & FID $(\downarrow)$\\
        \midrule
        Gaussian Perturbation & $27.20$ & $18.04$ & $28.18$ & $16.66$ & $29.20$ & $14.66$ & $30.09$ & $11.30$ & $33.90$ & $7.28$\\
        Latent Optimization & $28.63$ & $19.93$ & $29.81$ & $17.02$ & $29.84$ & $14.97$ & $30.23$ & $12.06$ & $33.52$ & $7.37$ \\
        Retrieval-Augmented & $30.31$ & $15.25$ & $29.93$ & $15.01$ & $30.26$ & $14.74$ & $30.55$ & $14.24$ & $32.09$ & $10.07$\\
        \midrule
        \shortname{}$_{\text{GAN}}$ & $\mathbf{32.16}$ & $\mathbf{10.26}$ & $\mathbf{32.24}$ & $\mathbf{9.96}$ & $\mathbf{32.65}$ & $\mathbf{9.63}$ & $\mathbf{32.63}$ & $\mathbf{9.22}$ & $\mathbf{34.16}$ & $\mathbf{6.78}$\\
        \bottomrule
    \end{tabular}
    }
\end{table*}

\begin{table*}[ht!]
    \centering
    \caption{Ablation studies. Image-text relevance on MS-COCO, evaluated by SOA-C ($\uparrow$) \& SOA-I ($\uparrow$).}
    \label{tab:ablation_2}    
    \scalebox{0.8}{
    \begin{tabular}{lcccccccccc}
    \toprule
         &  \multicolumn{2}{c}{$0\%$} & \multicolumn{2}{c}{$0.1\%$} & \multicolumn{2}{c}{$1\%$} & \multicolumn{2}{c}{$10\%$} & \multicolumn{2}{c}{$100\%$}\\
         Component & SOA-C &  SOA-I  & SOA-C &  SOA-I  & SOA-C&  SOA-I & SOA-C &  SOA-I  & SOA-C &  SOA-I\\
        \midrule
        Gaussian Perturbation & $36.84$ & $54.16$ & $38.84$ & $56.24$ & $41.86$ & $59.48$ & $47.94$ & $64.64$ & $60.31$ & $74.48$\\
        Latent Optimization & $41.74$ & $59.53$ & $44.55$ & $62.07$ & $45.89$ & $63.02$ & $48.85$ & $65.98$ & $61.32$ & $\mathbf{74.88}$\\
        Retrieval-Augmented & $54.95$ & $69.59$ & $54.74$ & $69.35$ & $\mathbf{55.43}$ & $69.98$ & $55.08$ & $69.93$ & $57.98$ & $72.61$ \\
        \midrule
        \shortname{}$_{\text{GAN}}$ & $\mathbf{55.99}$ & $\mathbf{71.25}$ & $\mathbf{55.02}$ & $\mathbf{70.16}$ & $55.34$ & $\mathbf{70.59}$ & $\mathbf{55.34}$ & $\mathbf{71.35}$ & $\mathbf{61.30}$ & $74.84$\\
        \bottomrule
    \end{tabular}
    }
\end{table*}
\begin{figure}[t!]
    \centering
    \includegraphics[width=0.7\linewidth]{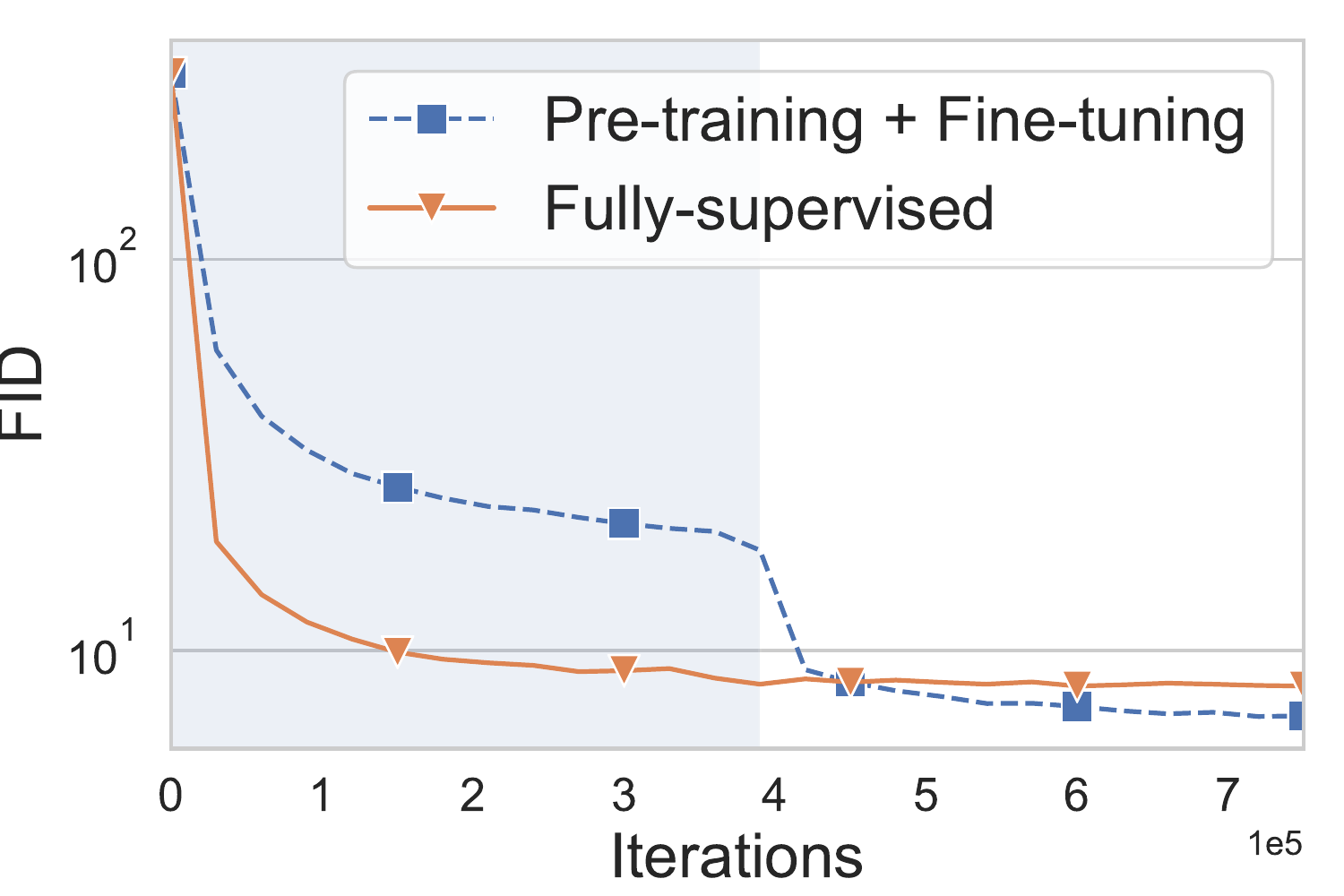}
    \caption{Fine-tuning after unsupervised pre-training (blue background) leads to better performance than fully-supervised training from scratch.}
    \label{fig:fid_is_curve}
\end{figure}

\paragraph{Effectiveness of Latent Optimization.}
We verify the proposed contrastive latent optimization (CLO) component by conducting an experiment on MS-COCO~\cite{lin2014microsoft}. For simplicity, we do not use any image feature to generate pseudo text features. Instead, text features are initialized uniformly on the hyper-sphere by $\hb_{ij} = \epsilonb_{ij}/\Vert \epsilonb_{ij} \Vert_2, \epsilonb_{ij} \sim \mathcal{N}(\mathbf{0}, \mathbf{I})$, which are then updated using \eqref{eq:sampling}. We train two \lafitegan{} models for 800K iterations with a batch size of 64. The fully-supervised model is trained with ground-truth image-text pairs from scratch. The other model is first pre-trained with pseudo image-text pairs generated by CLO, then fine-tuned with ground-truth pairs. As we can see in Figure \ref{fig:fid_is_curve}, pre-training with CLO leads to better results than training from scratch.

\paragraph{Component Analysis.}
We further pre-train two models with pseudo text features generated by Gaussian perturbation, CLO and retrieval-augmented method respectively. We then fine-tune all three pre-trained models on 0$\%$, 0.1$\%$, 1$\%$, 10$\%$, 100$\%$ of ground-truth image-text pairs from MS-COCO respectively. The  results are provided in Table \ref{tab:ablation_1}
and Table \ref{tab:ablation_2}. 
The proposed retrieval method leads to the best results in terms of both generation quality and image-text correspondence in fewer-shot settings. However, it obtains worse results than Gaussian perturbation and CLO when more captions are provided. This happens because the model tend to over-fit the specific structure of templates. 
CLO obtains much better image-text correspondence compared to Gaussian perturbation, which verifies our previous motivation and assumption. Finally, combining all three components together leads to the best results overall, and it can reduce the over-fitting problem.

\section{Conclusion}
In this work, we have presented \shortname{}, a novel unsupervised pre-training for text-to-image generation task which only requires image samples while obtains good image-text correspondence and generation quality. \shortname{} has good transferability and can benefit many different settings. 
The proposed method obtains promising results in the fully-supervised setting across different datasets. We also evaluated the proposed method on few-shot and semi-supervised and obtains better results compared to related methods. We believe that such an unsupervised pre-training method will be beneficial to the area because of its promising performance under different settings and low need of human workload in constructing image-text dataset.

\paragraph{Acknowledgement}
This work is partially supported by NSF CISE-2223292, NSF IIS-1910492, an Amazon research award, and an Adobe gift fund.

\clearpage
{\small
\bibliographystyle{ieee_fullname}
\bibliography{egbib}
}

\clearpage

\appendix
\onecolumn

\section{Proof of the Theorem}
\textbf{Theorem \ref{thm:gradient_bound}} 
\textit{
    Let $\{\xb_j^\prime\}_{i=1}^n$ be a mini-batch of generated images, $\{\hb_i\}_{i=1}^n$ be corresponding text features which are fed into the generator $G_{\thetab}$. For the contrastive loss $\mathcal{L}$ in \eqref{eq:contrastive_loss}, we can prove that
    \begin{align*}
        \Vert \nabla_{\thetab} \mathcal{L}\Vert_2 \leq na + n^2a\sigma.
    \end{align*}
    $a\geq 0$ is a constant related to the CLIP image encoder and generator, $\sigma \geq 0$ denotes the standard deviation of $\{c_{ji}\}$ where 
    \begin{align*}
        c_{ji} = \exp(\text{Sim}(f_{\text{img}}(\xb_j^\prime), \hb_i)/\tau)/\sum_{k=1}^n \exp(\text{Sim}(f_{\text{img}}(\xb_k^\prime), \hb_i)/\tau)
    \end{align*}
}
\begin{proof}
    For clearness, we use 
    \[
    s_{ji} = \text{Sim}(f_{\text{img}}(\xb_j^\prime), \hb_i)/\tau
    \]
    \begin{align*}
    \nabla_{\thetab} \mathcal{L} = & \nabla_{\thetab} \sum_{i=1}^n \log \dfrac{\exp(s_{ii})}{\sum_{j=1}^n \exp(s_{ji})} \\
    = &  \sum_{i=1}^n \dfrac{\sum_{k=1}^n \exp(s_{ki})}{\exp(s_{ii})} \dfrac{\sum_{j=1}^n \exp(s_{ji}) \nabla_{\thetab} \exp(s_{ii}) - \exp(s_{ii}) \nabla_{\thetab} \sum_{j=1}^n \exp(s_{ji})}{\{\sum_{j=1}^n \exp(s_{ji})\}^2} \\
    = & \sum_{i=1}^n \dfrac{1}{\exp(s_{ii})\sum_{k=1}^n \exp(s_{ki})} \{\sum_{j=1}^n \exp(s_{ji}) \exp(s_{ii})  \nabla_{\thetab} s_{ii} - \exp(s_{ii}) \sum_{j=1}^n \exp(s_{ji})\nabla_{\thetab} s_{ji}  \}\\
    = & \sum _{i,j}^n\dfrac{\exp(s_{ji})}{\sum_{k}^n \exp(s_{ki})} \nabla_{\thetab} s_{ii} - \dfrac{\exp(s_{ji})}{\sum_{k=1}^n \exp(s_{ki})} \nabla_{\thetab} s_{ji}\\
    = & \sum_{i,j}^n c_{ji} \nabla_{\thetab} s_{ii} - c_{ji}  \nabla_{\thetab} s_{ji} \\
    = &  \sum_{i,j}^n (1/n - 1/n + c_{ji})(\nabla_{\thetab} s_{ii} - \nabla_{\thetab} s_{ji})\\
    = & \sum_{i,j}^n d_{ij}/n+ \sum_{i,j}^n ( c_{ji} - 1/n)d_{ij}
    \end{align*}
    where $d_{ij} = \nabla_{\thetab} s_{ii} - \nabla_{\thetab} s_{ji}$. Let $a = 2 \max \Vert \nabla_{\thetab} s_{i,j} \Vert_2$, which is a constant only depends on pre-trained CLIP image encoder and the generator, then $\Vert d_{ij}\Vert_2 = \Vert \nabla_{\thetab} s_{ii} - \nabla_{\thetab} s_{ji} \Vert_2 \leq a$. Then
    \begin{align*}
            \Vert \nabla_{\thetab} \mathcal{L} \Vert_2 \leq & \Vert \sum_{i,j}^n d_{ij}/n\Vert_2 + \Vert \sum_{i,j}^n ( c_{ji} - 1/n)d_{ij}\Vert_2 \\
            \leq & \dfrac{n^2 a}{n} + a \Vert  \sum_{i,j}^n ( c_{ji} - 1/n) \Vert_2 \\
            \leq & na + a\sqrt{\{\sum_{i,j}^n ( c_{ji} - 1/n)\}^2} \\
             = & na + n^2 a\sqrt{n^2 \sum_{i,j}^n ( c_{ji} - 1/n)^2/n^2} \\ 
             = & na + n^2a \sigma
    \end{align*}
\end{proof}

\section{Experiments}

\subsection{Settings}
Some details of datasets are provided in Table \ref{tab:exp_setting}. Note Birdsnap provides 50K URLs to images, while we find that only 40K of them are downloadable in May 2022.

To prevent overlapping between pre-training and downstream datasets in near-domain and multi-domain pre-training, we apply MD5~\cite{rivest1992md5} and SHA256~\cite{Gilbert2003SecurityAO} methods. We remove the images from the pre-training dataset when they have the same values with any image from the downstream dataset.

The hyper-parameter tuning is based on grid search. Specifically, the update step $T$ for latent optimization is selected from $\{0, 5, 10\}$, step size $\lambda$ is selected from $\{0.01, 0.02, 0.05, 0.1, 0.2, 0.5\}$. We use $T=10, \lambda=0.01$ while updating pseudo text features generated from retrieval-augmented method. We use $T=10, \lambda=0.3$ while pseudo updating features intialized by uniformly sampling from hyper-sphere Other hyper-parameters follows previous work~\cite{zhou2021lafite}.

To prevent over-fitting, we start from training with only pseudo features generated with Gaussian perturbation, then gradually introduce pseudo features generated with proposed method. Their ratio can either be a linear function or a step function. Similar results are obtained in both cases.

\begin{table}[t!]
    \centering
    \caption{The data statistics of pre-training and fine-tuning datasets used in various scenarios. Size of validation/testing sets are indicated in parenthesis if applicable, while they are never used during training. We ensure that all the pre-training images do not overlap with images downstream dataset for task-level transfer scenarios. }
    \label{tab:exp_setting}
    \scalebox{0.75}{
    \begin{tabular}{llll|lll}
    \toprule
         \multicolumn{4}{c|}{ Pre-training}  &  \multicolumn{3}{c}{Downstream} \\
           Dataset  & \# Images &  URL  &  License & Dataset & \# Image-Text Pairs & License\\
        \midrule
        \multicolumn{4}{l}{\bf I. Unsupervised Warmup (In-Domain Pre-training)} \\
        MS-COCO~\cite{lin2014microsoft} & 83K(40K) &  \href{https://cocodataset.org/\#download}{COCO} & Custom &  MS-COCO & 83K(40K) & Custom \\
        CUB~\cite{WahCUB_200_2011} &  9K(3K) & \href{https://data.caltech.edu/records/20098}{CUB-200-2011} & CC-BY &  CUB &  9K(3K) & CC-BY \\   
        LN-COCO~\cite{pont2020connecting} & 134K(8K) & \href{https://google.github.io/localized-narratives/}{Local Narratives} & CC-BY 4.0 & LN-COCO &  134K(8K) & CC-BY 4.0 \\   
        MM-CelebA-HQ~\cite{xia2021tedigan} & 24K(6K) & \href{https://github.com/IIGROUP/MM-CelebA-HQ-Dataset}{MM-CelebA-HQ} & Custom & MM-CelebA-HQ &  24K(6K) & Custom\\   
        \midrule
        \multicolumn{5}{l}{\bf II.  Task-level Transfer (Near-Domain Pre-training)}     \\
        COCO 2017 unlabeled set & 123K & 
        \href{https://cocodataset.org/\#download}{COCO}
         & Custom & MS-COCO &  83K(40K) & Custom \\
        Birdsnap~\cite{Berg_2014_CVPR} & 50K & \href{http://thomasberg.org/datasets/birdsnap/1.1/birdsnap.tgz}{Birdsnap}  & Unknown & CUB &  9K(3K) & CC-BY\\   
        COCO 2017 unlabeled set & 123K &  \href{https://cocodataset.org/\#download}{COCO} & Unknown  & LN-COCO &   134K(8K) & CC-BY 4.0\\   
        FFHQ~\cite{karras2019style} & 70K & \href{https://github.com/NVlabs/ffhq-dataset}{FFHQ} & CC BY-NC-SA 4.0   & MM-CelebA-HQ & 24K(6K) & Custom \\
        \midrule        
        \multicolumn{4}{l}{\bf III. Task-level Transfer (Multi-Domain Pre-training)}     \\
         \multirow{4}{*}{The union of 3 datasets} & & &  & MS-COCO &  83K(40K) & Custom\\
         & &  & & CUB &  9K(3K) & CC-BY\\   
         & &  & & LN-COCO &   134K(8K) & CC-BY 4.0\\   
         & &  & & MM-CelebA-HQ & 24K(6K) & Custom\\    
         \bottomrule
    \end{tabular}
    }
\end{table}

\subsection{Generated Examples}
We provide some generated examples of \lafitegan{} here.
\begin{figure}
    \centering
    \includegraphics[width=0.8\linewidth]{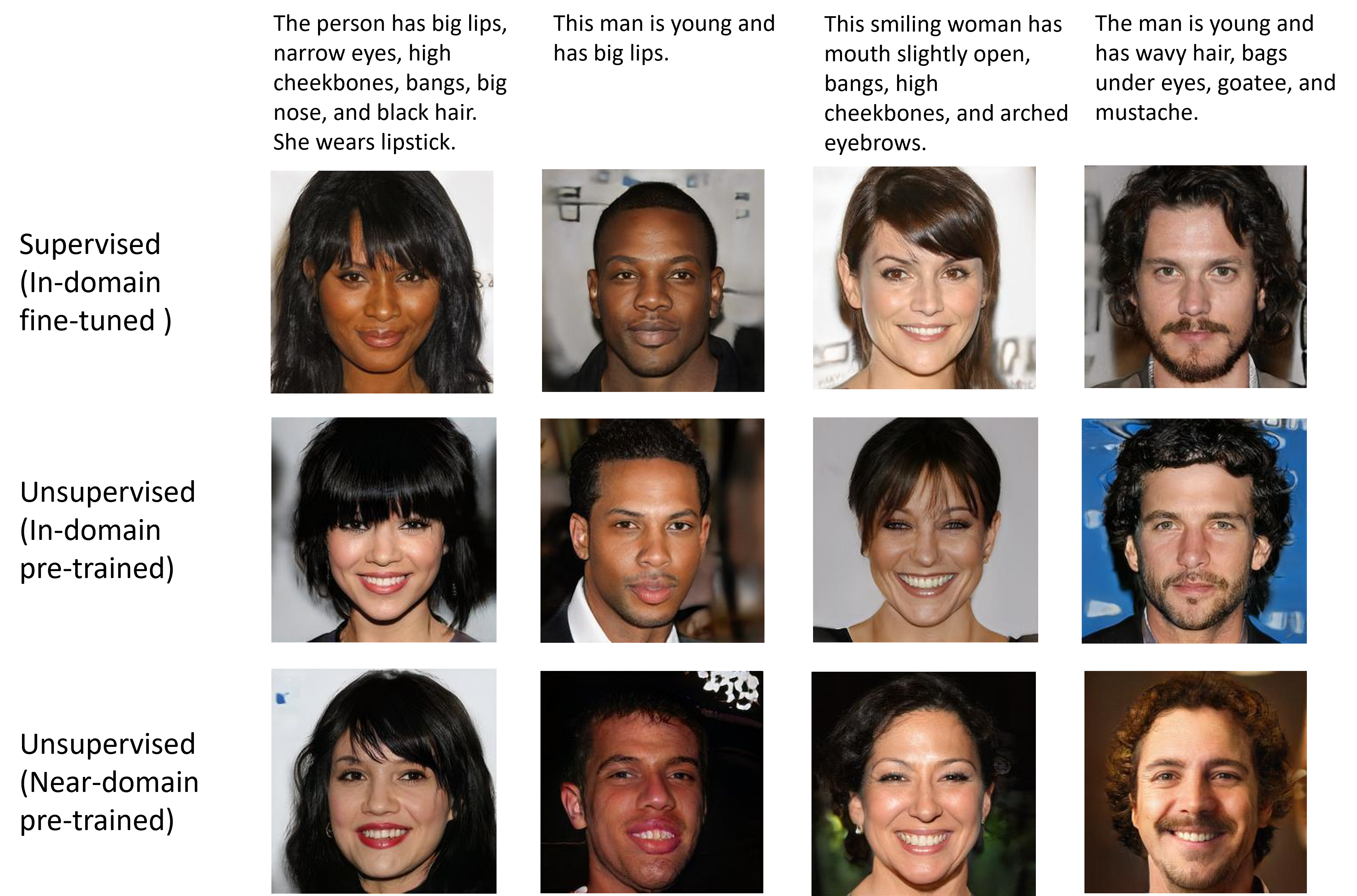}
    \caption{Generated examples on MM-CelebA-HQ dataset.}
    \label{fig:generated-celeba}
\end{figure}

\begin{figure}
    \centering
    \includegraphics[width=0.8\linewidth]{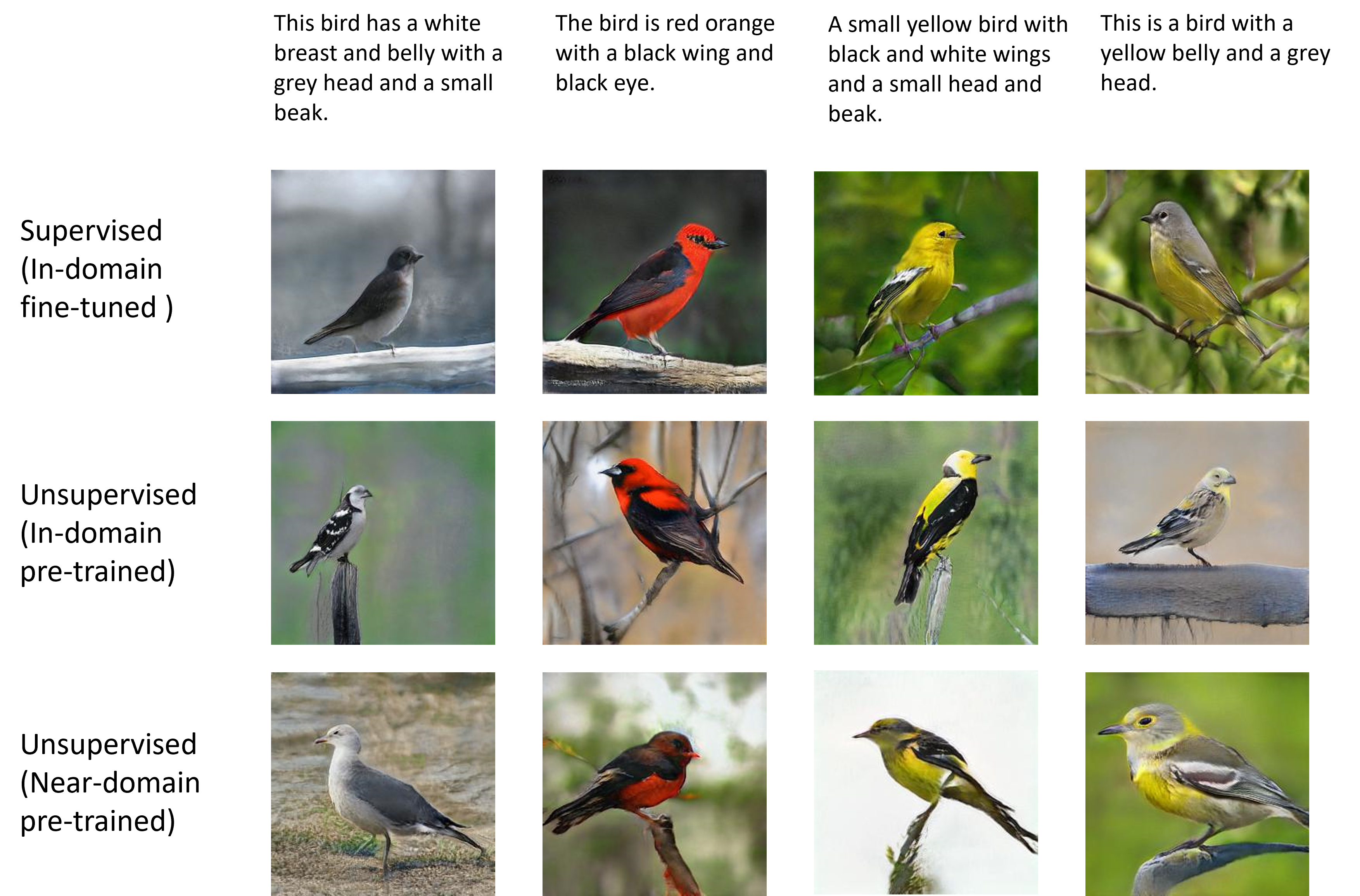}
    \caption{Generated examples on CUB dataset.}
    \label{fig:generated-cub}
\end{figure}

\begin{figure}
    \centering
    \includegraphics[width=0.8\linewidth]{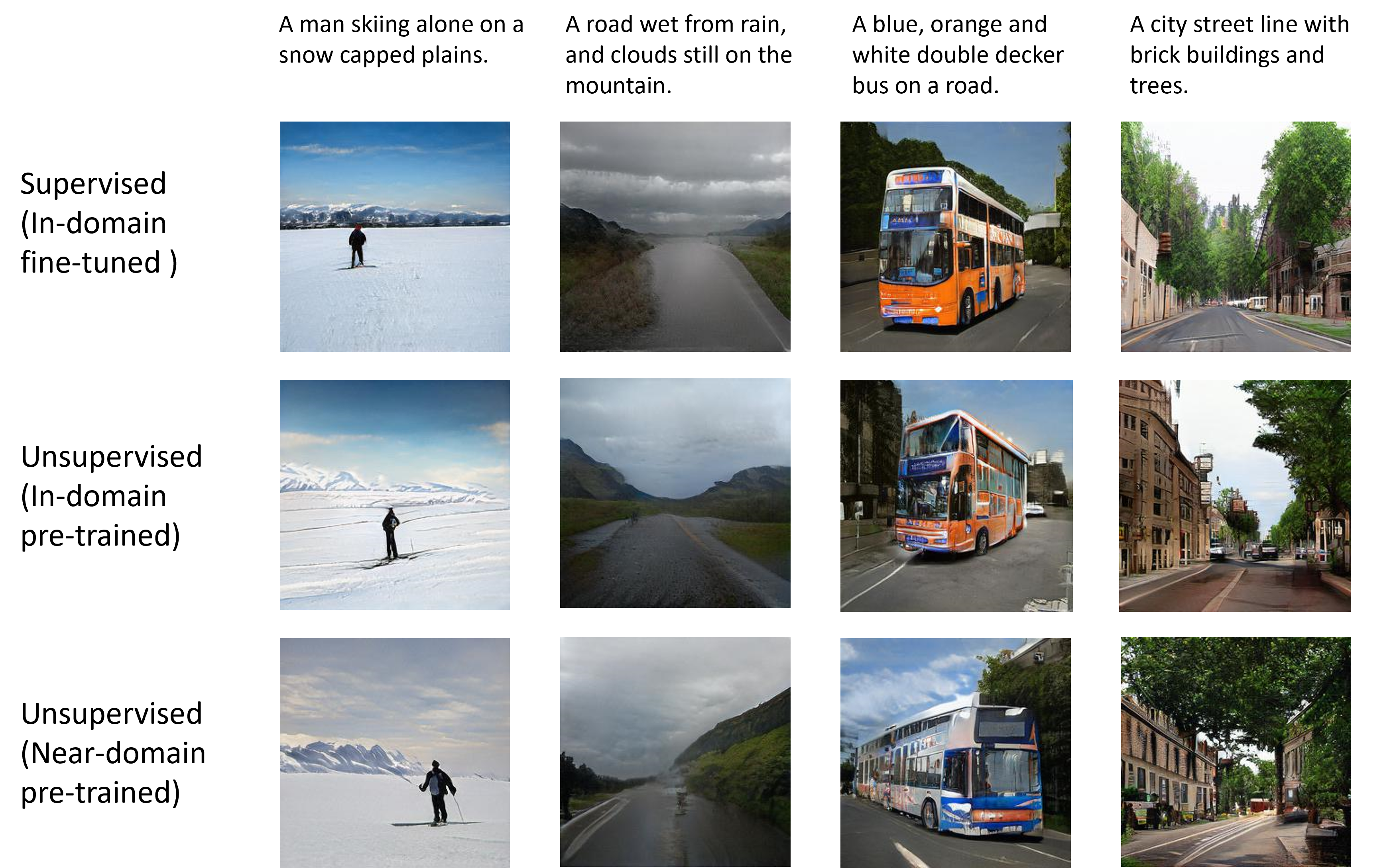}
    \caption{Generated examples on MS-COCO dataset.}
    \label{fig:generated-coco}
\end{figure}

\begin{figure}
    \centering
    \includegraphics[width=0.8\linewidth]{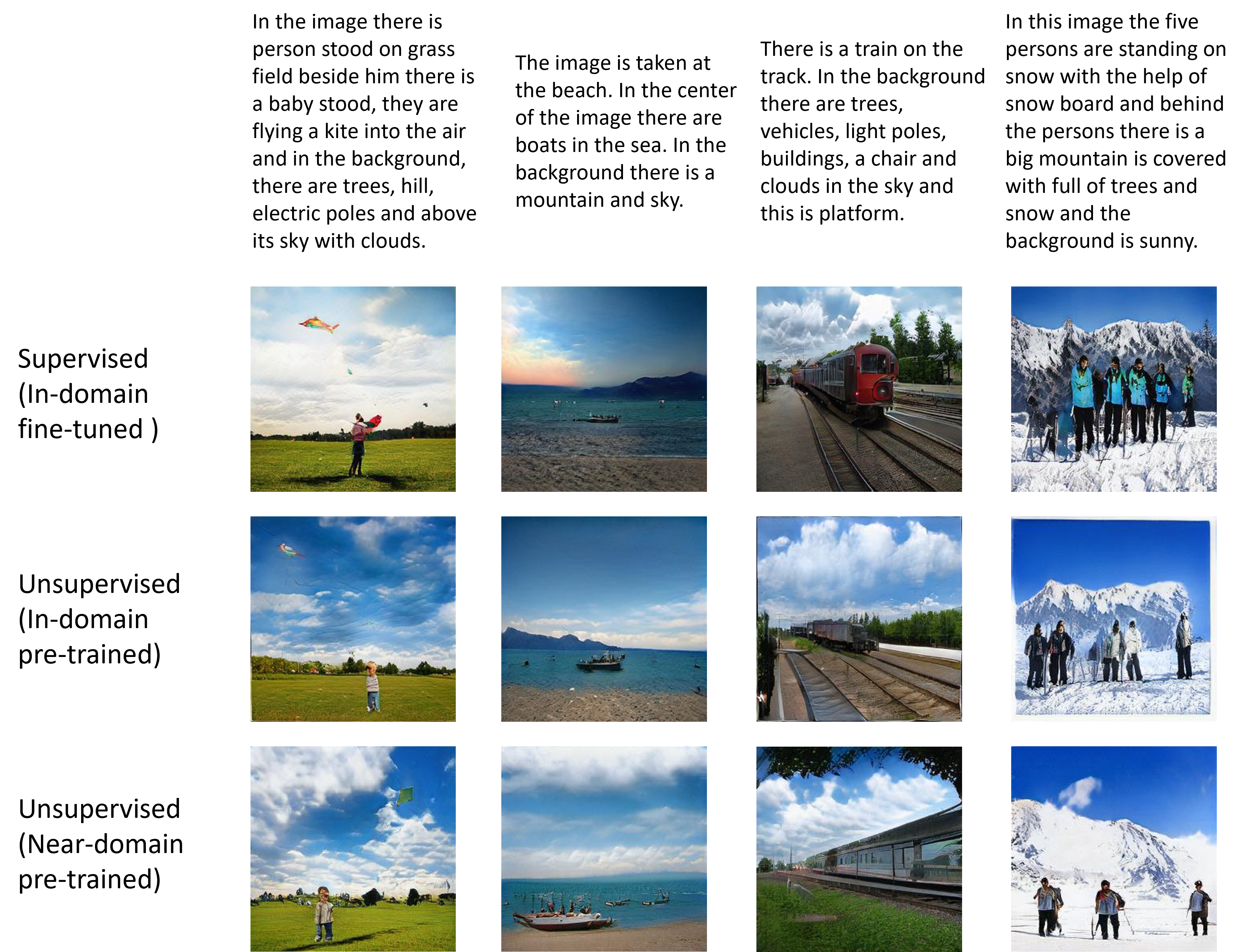}
    \caption{Generated examples on LN-COCO dataset.}
    \label{fig:generated-ln_coco}
\end{figure}

\end{document}

%% file: definitions.tex
\usepackage{amsmath,amsfonts,bm}

\newcommand{\RN}[1]{%
	\textup{\lowercase\expandafter{\it \romannumeral#1}}%
}

\newcommand{\Magenta}[1]{\color{magenta}{#1}\color{black}}

\ifx\note\undefined
\newcommand{\note}[1]{{\bf \Magenta{#1}}}
\else
\renewcommand{\note}[1]{{\bf \Magenta{#1}}}
\fi

\DeclareMathOperator{\hb}{\mathbf{h}}

\DeclareMathOperator{\wb}{\mathbf{w}}
\DeclareMathOperator{\xb}{\mathbf{x}}

 \DeclareMathOperator{\yb}{\mathbf{y}}
\DeclareMathOperator{\zb}{\mathbf{z}}

\newcommand{\thetab}{{\bm{\theta}}}

\newcommand{\epsilonb}{{\bm{\epsilon}}}

\ifx\BlackBox\undefined
\newcommand{\BlackBox}{\rule{1.5ex}{1.5ex}}  
\fi

\ifx\proof\undefined
\newenvironment{proof}{\par\noindent{\bf Proof\ }}{\hfill\BlackBox\\[2mm]}
\fi

\ifx\theorem\undefined
\newtheorem{theorem}{Theorem}
\fi
\ifx\example\undefined
\newtheorem{example}{Example}
\fi
\ifx\lemma\undefined
\newtheorem{lemma}[theorem]{Lemma}
\fi

\ifx\corollary\undefined
\newtheorem{corollary}[theorem]{Corollary}
\fi

\ifx\definition\undefined
\newtheorem{definition}[theorem]{Definition}
\fi

\ifx\proposition\undefined
\newtheorem{proposition}[theorem]{Proposition}
\fi

\ifx\remark\undefined
\newtheorem{remark}[theorem]{Remark}
\fi

\ifx\conjecture\undefined
\newtheorem{conjecture}[theorem]{Conjecture}
\fi

\ifx\factoid\undefined
\newtheorem{factoid}[theorem]{Fact}
\fi

\ifx\axiom\undefined
\newtheorem{axiom}[theorem]{Axiom}
\fi

\ifx\br\undefined
\newcommand{\br}{\color{red}}
\else
\renewcommand{\br}{\color{red}}
\fi

\ifx\bl\undefined
\newcommand{\bl}{\hspace*{0.8cm}}
\else
\renewcommand{\bl}{\hspace*{0.8cm}}
\fi